\documentclass{article}
\usepackage{times}

\usepackage[numbers]{natbib}  % numeric citations for preprint

\usepackage{imports}               % your cleaned-up package
\usepackage{bibentry}

\usepackage[margin=3cm]{geometry}

\makeatletter
\counterwithout{theorem}{section}

\newtoggle{draft}
\toggletrue{draft}

\newtoggle{nips}
\togglefalse{nips}

\newtoggle{aistats}
\togglefalse{aistats}

\let\regularcite\cite % Save the original \cite command
\renewcommand\cite[1]{
    \iftoggle{nips}{\parencite{#1}}{\regularcite{#1}} 
}

\DTMlangsetup[en-GB]{ord=raise,monthyearsep={,\space}}
\DTMsavedate{date}{2021-07-21}

\begin{document}

\mathtoolsset{showonlyrefs=false}  % This numbers all equations
 
\title{Islands of Stability: Tractable Lattice Structures for No-Regret Learning of Super-Stable Matchings}
\date{} % omit date

\author{%
    Larkin W. Liu\textsuperscript{1, 3,}\thanks{Corresponding author email: \texttt{larkin@riebaki.com}} \quad
    Zilong Wang\textsuperscript{2} \quad
    Jalal Etesami\textsuperscript{1} \quad
    Shuai Li\textsuperscript{2} \\
    \small
    \textsuperscript{1}Technische Universität München \quad
    \textsuperscript{2}Shanghai Jiao Tong University \quad
    \textsuperscript{3}Riebaki AI
}
% \footnotetext[4]{Corresponding author: \texttt{larkin@riebaki.com}}
\maketitle

\begin{abstract}
    The stable matching problem has long served as a foundational model for two-sided market design, yet classical solutions assume that complete preference orderings are known in advance. We study an online version of this setting, motivated by the challenges of learning a stable matching in a repeated game—specifically as it relates to super stability and efficient learning under partial information. Our model is designed to operate directly on pairwise comparative feedback, making it practical in settings where agents observe only comparative outcomes. This forms a more natural representation of preference elicitation, obviating the need to assign ordinal scores to represent ordinal preferences using numerical scores (common in previous work). Operating on direct pairwise feedback, we first present a learning algorithm, $\mathsf{P2ETG}$, with provable sublinear regret guarantees with respect to achieving any stable matching. Next, as a crucial extension, we present $\mathsf{PrefLID}$, an algorithm which exploits the lattice structure of stable matchings and the concept of super-stability, which characterizes matchings that remain stable under every linear extension of the partial preferences. This concept connects common stable matchings among different refinements of a bipartite partial order into connected graphs (termed islands). We demonstrate that these islands reveal structural properties of the preference-refinement space that can be exploited for more efficient super-stable matching search over partial orders. This allows $\mathsf{PrefLID}$ to sample more efficiently, making only the pairwise comparisons needed to resolve the problem structure within a single island. We provide a tractable means to construct this island structure, enabling its use in an efficient online setting and demonstrating that it achieves no-regret performance while avoiding full preference resolution. We illustrate the applicability of our approach through a natural application to LLM routing, where agents must be matched to tasks in a streaming setting, and further provide corresponding empirical results to corroborate the theoretical properties.
    
\end{abstract}

\section{Introduction}
\label{sec:litreview}

How can a two-sided market learn a stable matching from a multitude of potential solutions when neither side knows its own preferences and the only available feedback is the outcome of a noisy comparison between two alternatives? One classical answer, built on the competing-bandits formulation \citep{liu2020competing}, couples preference learning across agents through competition for partners; subsequent work has extended the programme to decentralized markets \citep{liu:21_decentralized_matching}, two-sided learning \citep{zhang2024decentralized}, many-to-one structures \citep{KongLi2024}, and sharper regret analysis \citep{kong:24_improved_analysis, LiWangKong2025}. This line of work, however, assumes each agent observes an \emph{absolute} stochastic reward for every potential partner. In many modern systems --- routing queries among large language models, matching users to services, assigning tasks to workers --- such a scalar is never observed. What is observed is comparative: this model answered better than that one, this worker was preferred to that one. Noisy pairwise comparisons of this kind are the currency of \emph{duelling bandits} and pure-exploration rank identification \citep{SuiEtAl2018, BusaFeketeEtAl2014, ChenEtAl2020}, and most recently of matching markets with relative feedback \citep{KongEtAl2026}, which identifies the player-optimal stable matching with sample complexity governed by the smallest gap among \emph{all} comparisons needed to recover the full ranking. Across all of these formulations one assumption persists: the learner must resolve \emph{every} agent's complete ranking before a stable matching can be certified.

The assumption is expensive, and it is also unnecessary. Resolving a pairwise comparison to high confidence costs samples inversely proportional to its preference gap, so recovering a complete ranking is bottlenecked by the single hardest comparison in the market --- even when that comparison is irrelevant to the matching ultimately chosen. Combinatorial work has long exploited this slack. The theory of stable marriage with incomplete information defines weak, strong, and \emph{super-stability} for partially specified preferences \citep{irving:1994_super_stable}, and possible and certain stability refine partial orderings into strict preferences \citep{AzizEtAl2020, AzizEtAl2022}. Preference elicitation shows that a stable matching can often be determined without eliciting every agent's full ranking \citep{DrummondBoutilier2014}, and matchings robust to preference perturbations remain stable across entire families of profiles \citep{mai:2018_finding_sm, gangam:2026_robust_sm}. The statistical translation is direct: if the observations induce a set of rankings that contains the true ranking with high probability, and a candidate matching is stable under \emph{every} ranking in that set, then the matching is stable with high probability. A purely deterministic check of robust stability thereby becomes a sound statistical stopping rule, and the learning goal shifts from resolving complete rankings to resolving the \emph{sufficient} partial rankings that certify a stable outcome.

This paper develops that observation into an online learning framework for two-sided matching under pairwise uncertainty. The learner maintains the uncertainty explicitly as a partial ordering, resolves comparisons adaptively, and stops as soon as the induced lattice of candidate stable matchings certifies a single outcome --- without requiring any agent's full ranking. The framework is naturally suited to AI systems that route queries among multiple large language models, where preferences over models are learned from noisy pairwise comparisons or user context feedback \citep{hu2024routerbench, liu:2026_online_moe}. Fully ranking every model for every query type is costly; focusing on the comparisons needed to certify a stable assignment lets the system stop exploring earlier with high-probability guarantees, while ensuring no user context or model has a clear incentive to switch.

\paragraph{Literature Comparison:} Table~\ref{tab:literature} summarizes the closest neighbouring research directions. The final row describes a target research formulation suggested by their intersection, rather than an established result.

\begin{table}[ht]
\centering \small \resizebox{\textwidth}{!}{
\begin{tabular}{lcccccc}
\toprule
Work &
Online &
Two-Sided Unknown &
Relative Feedback &
Partial Prefs. &
Stable Structure &
Primary Objective \\
\midrule
\cite{liu2020competing}
& \cmark & \xmark & \xmark & \xmark & Limited & Stable Regret \\

\cite{liu:21_decentralized_matching}
& \cmark & \xmark & \xmark & \xmark & Limited & Decentralized Regret \\

\cite{KongLi2023}
& \cmark & \xmark & \xmark & \xmark & Gale-Shapley & Player-Optimal Regret \\

\cite{kong:24_improved_analysis}
& \cmark & \xmark & \xmark & \xmark & \cmark & Improved Stable Regret \\

\cite{hosseini2024putting}
& \cmark & \xmark & \xmark & Partial & \cmark & Stable Identification \\

\cite{zhang2024decentralized}
& \cmark & \cmark & \xmark & \xmark & Gale-Shapley  & Two-Sided Learning \\

\cite{KongEtAl2026}
& \cmark & Partial & \cmark & Confidence-Based & Gale-Shapley  & Stable Identification \\

\cite{DrummondBoutilier2014}
& Elicitation & \cmark & Queries & \cmark & Stability-Based & Minimize Elicitation \\

\cite{AzizEtAl2020}
& \xmark & Uncertain & \xmark & \cmark & Super-Stability & Certain Stability \\

\cite{AzizEtAl2022}
& \xmark & Uncertain & Pairwise Probs. & \cmark & \xmark & Certain Stability \\

\cite{MaiVazirani2018}
& \xmark & Perturbations & \xmark & Profiles & \cmark & Robust Stability \\

\cite{GangamEtAl2026}
& \xmark & Changing Profiles & \xmark & Profiles & \cmark & Robust Stability \\

\cite{mirfakhar:26_relative_feedback}
& \cmark & \cmark & \cmark & Interviews (low-cost hints) & \cmark & Stable Identification \\

\cite{athanasopoulos:25_learning_equilibria}
& \cmark & \cmark & \xmark & \cmark & \cmark &  Stable Identification \\

\midrule
Our Formulation
& \cmark & \cmark & \cmark & \cmark & \cmark &
Stability Certification with Selective Learning \\
\bottomrule
\end{tabular}}
\caption{Schematic comparison of research most closely related to learning stable matchings under pairwise preference uncertainty.}
\label{tab:literature}
\end{table}

Table~\ref{tab:literature} shows that while each dimension has been studied in isolation, no online stable matching work combines \emph{relative feedback}, explicit \emph{partial preferences}, and \emph{stable structures} to achieve stability certification with selective sampling. Relative (Plackett--Luce) feedback is not new: \citet{KongEtAl2026} use it on proposer subsets, resolving each arm's top choice along the Gale--Shapley path (OGS). Our contribution is the learning problem and its certificate, not the feedback model: we learn \emph{both} sides, choose comparison pairs adaptively, and certify \emph{any} stable matching as super-stable while comparisons remain unresolved, with corresponding sample-complexity guarantees. Our algorithmic contributions in this work to the online stable matching literature are primarily:

\begin{tcbitemize}[
    raster columns=1,
    raster equal height=rows,
    raster row skip=4pt,
    colframe=indigo,
    boxrule=0.6pt,
    arc=3pt,
    left=6pt, right=6pt, top=4pt, bottom=4pt
]
    \tcbitem[colback=boxbg]
        \textbf{Adaptive Pairwise Feedback on Both Sides:} Building on the Plackett--Luce (pairwise Bradley--Terry) feedback model of \citet{KongEtAl2026}, we introduce an algorithm that learns \emph{both} sides of the market and adaptively selects which pairwise comparisons to resolve based on their relevance to stability certification.
    \tcbitem[colback=lightbg]
        \textbf{Dynamic Entry and Exit:} The proposed architecture seamlessly integrates agents entering and leaving the system dynamically, operating effectively under heterogeneous information constraints. (Sec. \ref{sec:preflid}.)
    \tcbitem[colback=cardbg]
        \textbf{Structural Insights and Sample Complexity:} We reveal core structural properties of the stable matching search space, yielding a geometric framework that isolates critical sampling regions to substantially minimize sample complexity in online learning. (Sec. \ref{sec:structural_properties}.)
\end{tcbitemize}

\section{Preliminaries}

\begin{figure}[h!]
    \centering
    \tikzset{every picture/.style={line width=0.75pt}}

\begin{tikzpicture}[x=0.75pt,y=0.75pt,yscale=-1,xscale=1]

% Define standard arrow style for double-ended arrows
\tikzset{dimarrow/.style={<->, >=stealth, line width=0.75pt}}

% Grid configuration constants:
% Row 1 Y = 20, Row 2 Y = 38, Dots Y = 54, Row 3 Y = 70
% Left Block X columns: Labels=0, Col1=30, Col2=60, Col3=90, Col4=120, Col5=150
% Right Block X columns: Labels=180, Col1=210, Col2=240, Col3=270, Col4=300, Col5=330
% Vertical divider at X=165

% ==================== LEFT BLOCK ====================

% Row Labels
\draw (0,20) node [anchor=center, font=\scriptsize] {$a_{1} :$};
\draw (0,38) node [anchor=center, font=\scriptsize] {$a_{2} :$};
\draw (0,54) node [anchor=center, font=\scriptsize] {$\vdots$};
\draw (0,70) node [anchor=center, font=\scriptsize] {$a_{5} :$};

% Row 1: b1, b2, b3, b4, b5
\draw (30,20) node [anchor=center, font=\scriptsize] {$b_{1}$};
\draw (60,20) node [anchor=center, font=\scriptsize] {$b_{2}$};
\draw (90,20) node [anchor=center, font=\scriptsize] {$b_{3}$};
\draw (105,20) node [anchor=center, font=\scriptsize] {$\succ$};
\draw (120,20) node [anchor=center, font=\scriptsize] {$b_{4}$};
\draw (135,20) node [anchor=center, font=\scriptsize] {$\succ$};
\draw (150,20) node [anchor=center, font=\scriptsize] {$b_{5}$};

% Row 2: b1, b2, b3, b5, b4 with arrows
\draw (30,38) node [anchor=center, font=\scriptsize] {$b_{1}$};
\draw (45,38) node [anchor=center, font=\scriptsize] {$\succ$};
\draw (60,38) node [anchor=center, font=\scriptsize] {$b_{2}$};
\draw (90,38) node [anchor=center, font=\scriptsize] {$b_{3}$};
\draw (120,38) node [anchor=center, font=\scriptsize] {$b_{5}$};
\draw (135,38) node [anchor=center, font=\scriptsize] {$\succ$};
\draw (150,38) node [anchor=center, font=\scriptsize] {$b_{4}$};
\draw [dimarrow] (30,28) -- (90,28); % Arrow placed directly below row 2

% Row 3: b2, b1, b4, b5, b3 with arrows
\draw (30,70) node [anchor=center, font=\scriptsize] {$b_{2}$};
\draw (45,70) node [anchor=center, font=\scriptsize] {$\succ$};
\draw (60,70) node [anchor=center, font=\scriptsize] {$b_{1}$};
\draw (75,70) node [anchor=center, font=\scriptsize] {$\succ$};
\draw (90,70) node [anchor=center, font=\scriptsize] {$b_{4}$};
\draw (105,70) node [anchor=center, font=\scriptsize] {$\succ$};
\draw (120,70) node [anchor=center, font=\scriptsize] {$b_{5}$};
\draw (150,70) node [anchor=center, font=\scriptsize] {$b_{3}$};

\draw [dimarrow] (60,46) -- (120,46); % Reverted back above row 3
\draw [dimarrow] (120,77) -- (150,77); % Below row 3

% ==================== VERTICAL DIVIDER ====================
\draw (165,10) -- (165,80);

% ==================== RIGHT BLOCK ====================

% Row Labels
\draw (180,20) node [anchor=center, font=\scriptsize] {$b_{1} :$};
\draw (180,38) node [anchor=center, font=\scriptsize] {$b_{2} :$};
\draw (180,54) node [anchor=center, font=\scriptsize] {$\vdots$};
\draw (180,70) node [anchor=center, font=\scriptsize] {$b_{5} :$};

% Row 1: a1, a2, a3, a4, a5
\draw (210,20) node [anchor=center, font=\scriptsize] {$a_{1}$};
\draw (225,20) node [anchor=center, font=\scriptsize] {$\succ$};
\draw (240,20) node [anchor=center, font=\scriptsize] {$a_{2}$};
\draw (270,20) node [anchor=center, font=\scriptsize] {$a_{3}$};
\draw (300,20) node [anchor=center, font=\scriptsize] {$a_{4}$};
\draw (315,20) node [anchor=center, font=\scriptsize] {$\succ$};
\draw (330,20) node [anchor=center, font=\scriptsize] {$a_{5}$};

% Row 2: a1, a2, a3, a5, a4 with arrows
\draw (210,38) node [anchor=center, font=\scriptsize] {$a_{1}$};
\draw (225,38) node [anchor=center, font=\scriptsize] {$\succ$};
\draw (240,38) node [anchor=center, font=\scriptsize] {$a_{2}$};
\draw (270,38) node [anchor=center, font=\scriptsize] {$a_{3}$};
\draw (300,38) node [anchor=center, font=\scriptsize] {$a_{5}$};
\draw (315,38) node [anchor=center, font=\scriptsize] {$\succ$};
\draw (330,38) node [anchor=center, font=\scriptsize] {$a_{4}$};
\draw [dimarrow] (231,28) -- (304,28); % Arrow placed directly below row 2

% Row 3: a2, a1, a4, a5, a3 with arrows
\draw (210,70) node [anchor=center, font=\scriptsize] {$a_{2}$};
\draw (225,70) node [anchor=center, font=\scriptsize] {$\succ$};
\draw (240,70) node [anchor=center, font=\scriptsize] {$a_{1}$};
\draw (270,70) node [anchor=center, font=\scriptsize] {$a_{4}$};
\draw (285,70) node [anchor=center, font=\scriptsize] {$\succ$};
\draw (300,70) node [anchor=center, font=\scriptsize] {$a_{5}$};
\draw (315,70) node [anchor=center, font=\scriptsize] {$\succ$};
\draw (330,70) node [anchor=center, font=\scriptsize] {$a_{3}$};

\draw [dimarrow] (231,46) -- (304,46); % Reverted back above row 3
\draw [dimarrow] (233,77) -- (273,77); % Below row 3

\end{tikzpicture}
    \caption{\textbf{Bipartite Ranking Topology $\parOrder$:} We represent a bipartite partial ordering. Each agent $a \in \setA$ and $b \in \setB$ has an ordering over the other side. However, due to uncertainty, some of the preferences are not fully resolved, leaving sections of sequences unordered. For example, \(a_1\)'s partial ordering over the $b$'s has three cells, $\{b_1,b_2,b_3\}$, $\{b_4\}$, $\{b_5\}$, because \(b_1, b_2, b_3\) are mutually indifferent while \(\{b_1, b_2, b_3\} \succ b_4 \succ b_5\) is strictly ordered.} \label{fig:partial_orderings_and_cuts}
\end{figure}
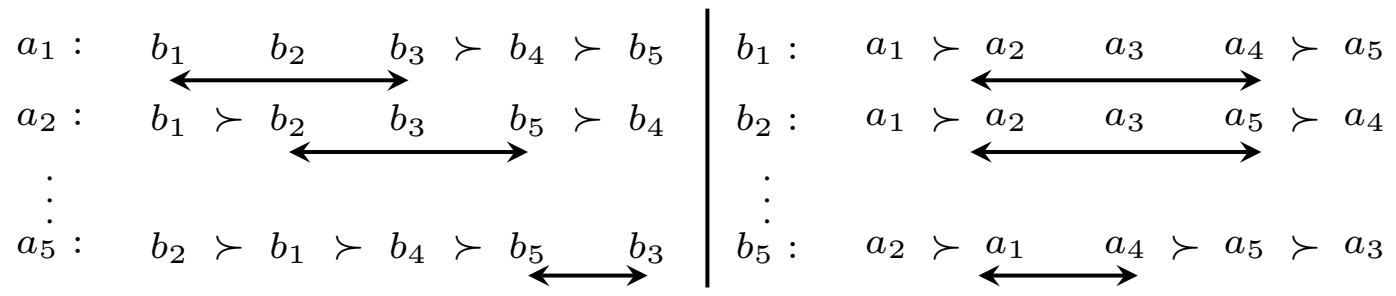

\paragraph{Stability \& Matching:} Let $\seT{A}$ and $\seT{B}$ be two finite, disjoint sets with $|\seT{A}| = N, \, |\seT{B}| = K $. Denote the set of all agents by $\seT{A} \cup \seT{B}$. Each agent $a \in \seT{A}$ truly possesses a strict, complete preference relation over the set $\seT{B}$, which we denote as $\succ_a$. For $b, b' \in \seT{B}$, we write $b \succ_a b'$ to indicate that $a$ strictly prefers $b$ to $b'$. Analogously, each agent $b \in \seT{B}$ has a strict, complete preference relation $\succ_b$ over $\seT{A}$. Completeness ensures that every agent ranks all members of the opposite set, and strictness precludes ties. We denote $\succ_a$ as a strict ordering for agent $a \in \setA$ over the set $\setB$, and  $\succ_b$ vice versa, for $N$ agents on the A-side and, $K$ agents on the B-side. 

\begin{definition}[Partial Orderings \& Bipartite Ranking Topology]\label{def:bipartite_partial_ordering}
 A \textbf{bipartite partial ordering} on $(\seT{A},\seT{B})$ is a pair $(\succeq_A,\succeq_B)$ with $\succeq_A\subseteq\seT{A}\times\seT{B}$ and $\succeq_B\subseteq\seT{B}\times\seT{A}$, where $\succeq_a$ partially orders $\seT{B}, \forall a\in\seT{A}$) and $\succeq_b$ partially orders $\seT{A}, \forall b\in\seT{B}$. Each partial order $\succeq_a$ induces a partition of $\seT{B}$ into cells of $\succeq_a$-indifferent agents; a strict, complete refinement of the partial order is a linear extension (a total order) on these cells. The \textbf{bipartite ranking topology} $\parOrder$ is the set of such strict, complete refinements $\succ^{(c)}_{AB} \in \parOrder$. (see Fig. \ref{fig:partial_orderings_and_cuts})
\end{definition}

\begin{definition}[Matching]
    A \textbf{matching} is a bijection $\matcH: \seT{A} \to \seT{B} $ (or $\matcH: \seT{B} \to \seT{A} $) such that \textup{I)} For every $a \in \seT{A}$, $\matcH(a) \in \seT{B}$; \textup{II)} For every $b \in \seT{B}$, $\matcH(b) \in \seT{A}$; \textup{III)} For every $x \in \seT{A} \cup \seT{B} $, $\matcH(\matcH(x)) = x$. Equivalently, a matching is a set of $N$ disjoint pairs $(a, b)$ with $a \in \seT{A}$ and $b \in \seT{B}$ that partitions $ \seT{A} \cup \seT{B} $.
    % \begin{enumerate}
    %     \item For every $a \in \seT{A}$, $\matcH(a) \in \seT{B}$;
    %     \item For every $b \in \seT{B}$, $\matcH(b) \in \seT{A}$;
    %     \item For every $x \in \seT{A} \cup \seT{B} $, $\matcH(\matcH(x)) = x$.
% \end{enumerate}
\end{definition}

\begin{remark}[Uneven Matching]
    In practice $N \leq K$ to ensure each agent in $\setA$ is matched. Intuitively, a matching assigns each agent in $\seT{A}$ to a unique partner in $\seT{B}$, and vice versa. If $N = K$ them no agent remains unmatched.
\end{remark}

\paragraph{Stability:} The notion of stability captures the absence of mutually beneficial deviations. A pair of agents who are not matched to each other forms a \emph{blocking pair} if they would both prefer to be together over their current assignments.

\begin{definition}[Blocking Pair]
    Given a matching $\matcH$, an unordered pair $(a, b) \in \seT{A} \times \seT{B} $ with $\matcH(a) \neq b$ is a \textbf{blocking pair} if: I) $b \succ_a \matcH(a)$ (agent $a$ prefers $b$ to his or her current match); and II) $a \succ_b \matcH(b)$ (agent $b$ prefers $a$ to his or her current match).
    % \begin{enumerate}
    %     \item $b \succ_a \matcH(a)$ (agent $a$ prefers $b$ to his or her current match);
    %     \item $a \succ_b \matcH(b)$ (agent $b$ prefers $a$ to his or her current match).
% \end{enumerate}
\end{definition}

\begin{definition}[Stable Matching]
$\matcH \in \SMSet$ denotes a set of stable matchings. $\matcH$ is \textbf{stable} if there exists no blocking pairs. Equivalently, for every $(a, b) \in A \times B$ with $\matcH(a) \neq b$,
$
\neg \bigl( b \succ_a \matcH(a) \ \text{and} \ a \succ_b \matcH(b) \bigr).
$
\end{definition}

\paragraph{Existence of Stable Matchings:}
A fundamental result due to \citep{Gale:1962} is that a stable matching always exists in any bipartite set of agents. The proof is constructive and given by the Gale-Shapley algorithm, also known as deferred acceptance. In the version where agents in $A$ propose, the Gale-Shapley (GS) algorithm illustrated in Alg. \ref{alg:gale_shapley} finds at least one stable matching which happens to be the proposer optimal, receiver pessimal stable matching. The algorithm terminates after at most $|\setA| \cdot |\setB|$ proposals and produces a stable matching\footnote{In our case we may be left with $\max\{ N-K, K-N \}$ unmatched agents if $N \ne K$.}. Moreover, it yields the matching that is optimal for the proposing side: each agent in $\setA$ receives the best possible partner among all stable matchings, while each agent in $\setB$ receives the worst possible partner among all stable matchings.

\subsection{Matchings on Partial Orderings}

We denote by \ textit{bipartite ranking topology} $\parOrder$, as the set of strict, complete refinements (linear extensions) of a single bipartite partial ordering, with members indexed by $c$, such that $\succ^{(c)}_{AB} \in \parOrder$ is a fully-resolved total order. The underlying partial ordering induces a partition of each side into cells of mutually indifferent agents. Each refinement is a total order on those cells. Between the strongest and weakest notions of stability under such ties lie several intermediate concepts, each balancing permissiveness with robustness against alternative tie-breakings.

\begin{definition}[Super-Stability] \label{def:super_stability}
     A pair $(a,b)$ is a \emph{super-blocking pair} if $a$ and $b$ are unmatched to each other and each prefers the other to their partner or is indifferent between them. $\matcH$ is \textbf{super-stable} if no super-blocking pairs exist. 
\end{definition}

Super-stability (Def. \ref{def:super_stability}) is the strongest of these notions: a matching remains stable under every refinement of the ties into strict orders. For any bipartite partially ordered instance $\parOrder$, a super-stable matching $\matcH^*$ need not exist; however \citep{irving:1994_super_stable} provided a polynomial-time algorithm that either finds such an $\matcH^*$ or determines that none exists.

\paragraph{Weak \& Strong Stability:} In the domain of bipartite partial orderings, finer concepts of stability emerge. Weak stability permits blocking only when both agents strictly gain, ensuring existence via tie-breaking but allowing indifference-based blocks; strong stability forbids any pair where at least one strictly benefits and the other does not lose, yielding a distributive lattice when existent. Together with super-stability, these notions form the strict hierarchy super-stability $\Rightarrow$ strong stability $\Rightarrow$ weak stability. Super-stability (Def. \ref{def:super_stability}) guarantees stability under the broadest interpretation of indifference. By contrast, strong stability requires stability for \emph{some} refinement of a partial ordering, while weak stability may fail for refinements that resolve ties unfavourably.

\section{Structural Properties \& Algorithms} \label{sec:structural_properties}

\begin{figure}[h!]
    \centering
    \begin{subfigure}[b]{0.48\linewidth}
        \centering
        \tikzset{every picture/.style={line width=0.75pt}} %set default line width to 0.75pt        
\begin{tikzpicture}[x=0.75pt,y=0.75pt,yscale=-0.59,xscale=0.59]
%uncomment if require: \path (0,428); %set diagram left start at 0, and has height of 428

% ===== PRIMARY COLOUR DEFINITIONS =====
% slate-navy: rgb(26,54,93)
% Nature green: rgb(56,161,105)

%Shape: Circle [id:dp27506112003243066]  (ρ1*)
\draw [color={rgb, 255:red, 26; green, 54; blue, 93},
       fill={rgb, 255:red, 26; green, 54; blue, 93}, fill opacity=0.08]
      (110.57,55.64) .. controls (110.57,48.98) and (115.97,43.58) .. (122.63,43.58)
      .. controls (129.3,43.58) and (134.7,48.98) .. (134.7,55.64)
      .. controls (134.7,62.3) and (129.3,67.71) .. (122.63,67.71)
      .. controls (115.97,67.71) and (110.57,62.3) .. (110.57,55.64) -- cycle ;

%Straight Lines
\draw [color={rgb, 255:red, 26; green, 54; blue, 93}] (35.42,13.54) -- (120.87,43.58) ;
\draw [color={rgb, 255:red, 26; green, 54; blue, 93}] (36.55,99.12) -- (115.34,65.95) ;
\draw [color={rgb, 255:red, 26; green, 54; blue, 93}] (124.77,43.33) -- (218.27,13.17) ;
\draw [color={rgb, 255:red, 26; green, 54; blue, 93}] (93.86,156.05) -- (116.48,65.95) ;
\draw [color={rgb, 255:red, 26; green, 54; blue, 93}] (175.04,155.04) -- (131.93,65.19) ;
\draw [color={rgb, 255:red, 26; green, 54; blue, 93}] (217.13,98.75) -- (130.8,64.44) ;

%Shape: Ellipse (L1^(1))
\draw [color={rgb, 255:red, 26; green, 54; blue, 93},
       fill={rgb, 255:red, 26; green, 54; blue, 93}, fill opacity=0.06]
      (18.83,56.33) .. controls (18.83,32.7) and (25.78,13.54) .. (34.35,13.54)
      .. controls (42.92,13.54) and (49.87,32.7) .. (49.87,56.33)
      .. controls (49.87,79.96) and (42.92,99.12) .. (34.35,99.12)
      .. controls (25.78,99.12) and (18.83,79.96) .. (18.83,56.33) -- cycle ;

%Shape: Ellipse (L1^(3))
\draw [color={rgb, 255:red, 26; green, 54; blue, 93},
       fill={rgb, 255:red, 26; green, 54; blue, 93}, fill opacity=0.06]
      (202.43,55.96) .. controls (202.43,32.32) and (209.38,13.17) .. (217.95,13.17)
      .. controls (226.52,13.17) and (233.47,32.32) .. (233.47,55.96)
      .. controls (233.47,79.59) and (226.52,98.75) .. (217.95,98.75)
      .. controls (209.38,98.75) and (202.43,79.59) .. (202.43,55.96) -- cycle ;

%Shape: Ellipse (L1^(2))
\draw [color={rgb, 255:red, 26; green, 54; blue, 93},
       fill={rgb, 255:red, 26; green, 54; blue, 93}, fill opacity=0.06]
      (93.6,157.43) .. controls (93.6,149.45) and (112,142.98) .. (134.7,142.98)
      .. controls (157.39,142.98) and (175.79,149.45) .. (175.79,157.43)
      .. controls (175.79,165.41) and (157.39,171.88) .. (134.7,171.88)
      .. controls (112,171.88) and (93.6,165.41) .. (93.6,157.43) -- cycle ;

%Shape: Circle (ρ2*)
\draw [color={rgb, 255:red, 26; green, 54; blue, 93},
       fill={rgb, 255:red, 26; green, 54; blue, 93}, fill opacity=0.08]
      (161.62,249.85) .. controls (161.62,243.19) and (167.03,237.79) .. (173.69,237.79)
      .. controls (180.35,237.79) and (185.75,243.19) .. (185.75,249.85)
      .. controls (185.75,256.52) and (180.35,261.92) .. (173.69,261.92)
      .. controls (167.03,261.92) and (161.62,256.52) .. (161.62,249.85) -- cycle ;

%Straight Lines (bottom cluster)
\draw [color={rgb, 255:red, 26; green, 54; blue, 93}] (86.48,207.75) -- (171.93,237.79) ;
\draw [color={rgb, 255:red, 26; green, 54; blue, 93}] (87.61,293.33) -- (166.4,260.16) ;
\draw [color={rgb, 255:red, 26; green, 54; blue, 93}] (175.82,237.54) -- (269.32,207.38) ;
\draw [color={rgb, 255:red, 26; green, 54; blue, 93}] (268.19,292.96) -- (181.86,258.65) ;

%Shape: Ellipse (L2^(1))
\draw [color={rgb, 255:red, 26; green, 54; blue, 93},
       fill={rgb, 255:red, 26; green, 54; blue, 93}, fill opacity=0.06]
      (69.89,250.54) .. controls (69.89,226.91) and (76.84,207.75) .. (85.41,207.75)
      .. controls (93.98,207.75) and (100.93,226.91) .. (100.93,250.54)
      .. controls (100.93,274.18) and (93.98,293.33) .. (85.41,293.33)
      .. controls (76.84,293.33) and (69.89,274.18) .. (69.89,250.54) -- cycle ;

%Shape: Ellipse (L2^(2))
\draw [color={rgb, 255:red, 26; green, 54; blue, 93},
       fill={rgb, 255:red, 26; green, 54; blue, 93}, fill opacity=0.06]
      (253.49,250.17) .. controls (253.49,226.54) and (260.43,207.38) .. (269.01,207.38)
      .. controls (277.58,207.38) and (284.53,226.54) .. (284.53,250.17)
      .. controls (284.53,273.8) and (277.58,292.96) .. (269.01,292.96)
      .. controls (260.43,292.96) and (253.49,273.8) .. (253.49,250.17) -- cycle ;

%Shape: Circle (ρm*)
\draw [color={rgb, 255:red, 26; green, 54; blue, 93},
       fill={rgb, 255:red, 26; green, 54; blue, 93}, fill opacity=0.08]
      (364.57,131.64) .. controls (364.57,124.98) and (369.97,119.58) .. (376.63,119.58)
      .. controls (383.3,119.58) and (388.7,124.98) .. (388.7,131.64)
      .. controls (388.7,138.3) and (383.3,143.71) .. (376.63,143.71)
      .. controls (369.97,143.71) and (364.57,138.3) .. (364.57,131.64) -- cycle ;

%Straight Lines (right cluster)
\draw [color={rgb, 255:red, 26; green, 54; blue, 93}] (310,61.33) -- (374.87,119.58) ;
\draw [color={rgb, 255:red, 26; green, 54; blue, 93}] (277,137.33) -- (371,142.33) ;
\draw [color={rgb, 255:red, 26; green, 54; blue, 93}] (383,121.33) -- (459,54.33) ;
\draw [color={rgb, 255:red, 26; green, 54; blue, 93}] (347.86,232.05) -- (370.48,141.95) ;
\draw [color={rgb, 255:red, 26; green, 54; blue, 93}] (429.04,231.04) -- (387,140.33) ;
\draw [color={rgb, 255:red, 26; green, 54; blue, 93}] (497,137.33) -- (386,140.33) ;

%Shape: Ellipse (Lm^(2))
\draw [color={rgb, 255:red, 26; green, 54; blue, 93},
       fill={rgb, 255:red, 26; green, 54; blue, 93}, fill opacity=0.06]
      (276.22,91.92) .. controls (285.99,70.4) and (300.23,55.83) .. (308.04,59.37)
      .. controls (315.84,62.91) and (314.25,83.23) .. (304.49,104.75)
      .. controls (294.72,126.27) and (280.47,140.84) .. (272.67,137.3)
      .. controls (264.86,133.76) and (266.45,113.44) .. (276.22,91.92) -- cycle ;

%Shape: Ellipse (Lm^(i))
\draw [color={rgb, 255:red, 26; green, 54; blue, 93},
       fill={rgb, 255:red, 26; green, 54; blue, 93}, fill opacity=0.06]
      (462.11,95.25) .. controls (452.78,72.01) and (453.43,53.17) .. (463.54,53.17)
      .. controls (473.66,53.17) and (489.42,72.01) .. (498.75,95.25)
      .. controls (508.08,118.49) and (507.44,137.33) .. (497.32,137.33)
      .. controls (487.2,137.33) and (471.44,118.49) .. (462.11,95.25) -- cycle ;

%Shape: Ellipse (Lm^(1))
\draw [color={rgb, 255:red, 26; green, 54; blue, 93},
       fill={rgb, 255:red, 26; green, 54; blue, 93}, fill opacity=0.06]
      (347.6,233.43) .. controls (347.6,225.45) and (366,218.98) .. (388.7,218.98)
      .. controls (411.39,218.98) and (429.79,225.45) .. (429.79,233.43)
      .. controls (429.79,241.41) and (411.39,247.88) .. (388.7,247.88)
      .. controls (366,247.88) and (347.6,241.41) .. (347.6,233.43) -- cycle ;

% ===== DASHED HIGHLIGHT REGIONS (Nature green, soft texture) =====
\draw  [color={rgb, 255:red, 56; green, 161; blue, 105},
        fill={rgb, 255:red, 56; green, 161; blue, 105}, fill opacity=0.18,
        dash pattern={on 4.5pt off 4.5pt}]
      (22,12) .. controls (42,2) and (151.23,-3.4) .. (208,7.33)
      .. controls (264.77,18.07) and (256,26.33) .. (254,53.33)
      .. controls (252,80.33) and (191,180.67) .. (144,183.33)
      .. controls (97,186) and (48,145.33) .. (28,115.33)
      .. controls (8,85.33) and (2,22) .. (22,12) -- cycle ;

\draw  [color={rgb, 255:red, 56; green, 161; blue, 105},
        fill={rgb, 255:red, 56; green, 161; blue, 105}, fill opacity=0.18,
        dash pattern={on 4.5pt off 4.5pt}]
      (270,59) .. controls (290,49) and (411.23,36.6) .. (468,47.33)
      .. controls (524.77,58.07) and (530,93.33) .. (528,120.33)
      .. controls (526,147.33) and (450,261.67) .. (403,264.33)
      .. controls (356,267) and (301,213.33) .. (281,183.33)
      .. controls (261,153.33) and (250,69) .. (270,59) -- cycle ;

\draw  [color={rgb, 255:red, 56; green, 161; blue, 105},
        fill={rgb, 255:red, 56; green, 161; blue, 105}, fill opacity=0.18,
        dash pattern={on 4.5pt off 4.5pt}]
      (72,202.33) .. controls (92,192.33) and (198.23,186.6) .. (255,197.33)
      .. controls (311.77,208.07) and (301,237.33) .. (299,264.33)
      .. controls (297,291.33) and (266,310.67) .. (219,313.33)
      .. controls (172,316) and (84,314.33) .. (64,284.33)
      .. controls (44,254.33) and (52,212.33) .. (72,202.33) -- cycle ;

% ===== TEXT NODES (unchanged) =====
\draw (371,68) node [anchor=north west][inner sep=0.75pt]   [align=left] {...};
\draw (15.5,49) node [anchor=north west][inner sep=0.75pt]  [font=\tiny]  {$\mathcal{L}_{1}^{( 1)}$};
\draw (120,145) node [anchor=north west][inner sep=0.75pt]  [font=\tiny]  {$\mathcal{L}_{1}^{( 2)}$};
\draw (199,44) node [anchor=north west][inner sep=0.75pt]  [font=\tiny]  {$\mathcal{L}_{1}^{( 3)}$};
\draw (463,85) node [anchor=north west][inner sep=0.75pt]  [font=\tiny]  {$\mathcal{L}_{m}^{( i)}$};
\draw (374,223) node [anchor=north west][inner sep=0.75pt]  [font=\tiny]  {$\mathcal{L}_{m}^{( 1)}$};
\draw (273,88) node [anchor=north west][inner sep=0.75pt]  [font=\tiny]  {$\mathcal{L}_{m}^{( 2)}$};
\draw (68,242) node [anchor=north west][inner sep=0.75pt]  [font=\tiny]  {$\mathcal{L}_{2}^{( 1)}$};
\draw (252,240) node [anchor=north west][inner sep=0.75pt]  [font=\tiny]  {$\mathcal{L}_{2}^{( 2)}$};
\draw (112,46) node [anchor=north west][inner sep=0.75pt]  [font=\tiny]  {$\rho _{1}^{*}$};
\draw (164,241) node [anchor=north west][inner sep=0.75pt]  [font=\tiny]  {$\rho _{2}^{*}$};
\draw (364,119) node [anchor=north west][inner sep=0.75pt]  [font=\tiny]  {$\rho _{m}^{*}$};
\draw (51,102) node [anchor=north west][inner sep=0.75pt]  [font=\tiny]  {$\isL^{( 1)}$};
\draw (156,275) node [anchor=north west][inner sep=0.75pt]  [font=\tiny]  {$\isL^{( 2)}$};
\draw (309,172.67) node [anchor=north west][inner sep=0.75pt]  [font=\tiny]  {$\isL^{( m)}$};
\end{tikzpicture}
        \caption{\textbf{Illustration of Island Lattice Structure:} Via the blocking function $\blockFunc$, a set of lattices $\setlaT$ can be mapped into a set of islands $\isL$, where within each island $\isL^{(m)}$ a common centroidal matching $\matcH^*_m$ is shared between the lattices $\laT \in \setlaT$.}
        \label{fig:islands}
    \end{subfigure}
    \hfill
    \begin{subfigure}[b]{0.48\linewidth}
        \centering
        \input{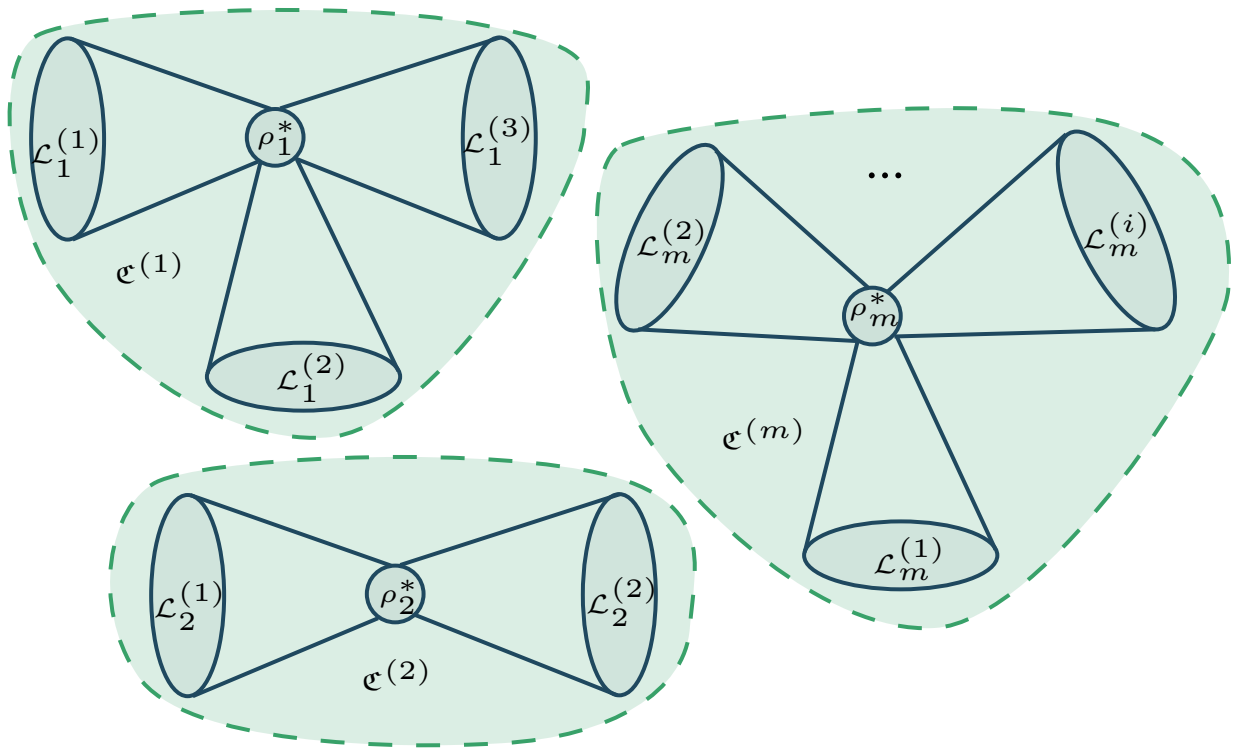}
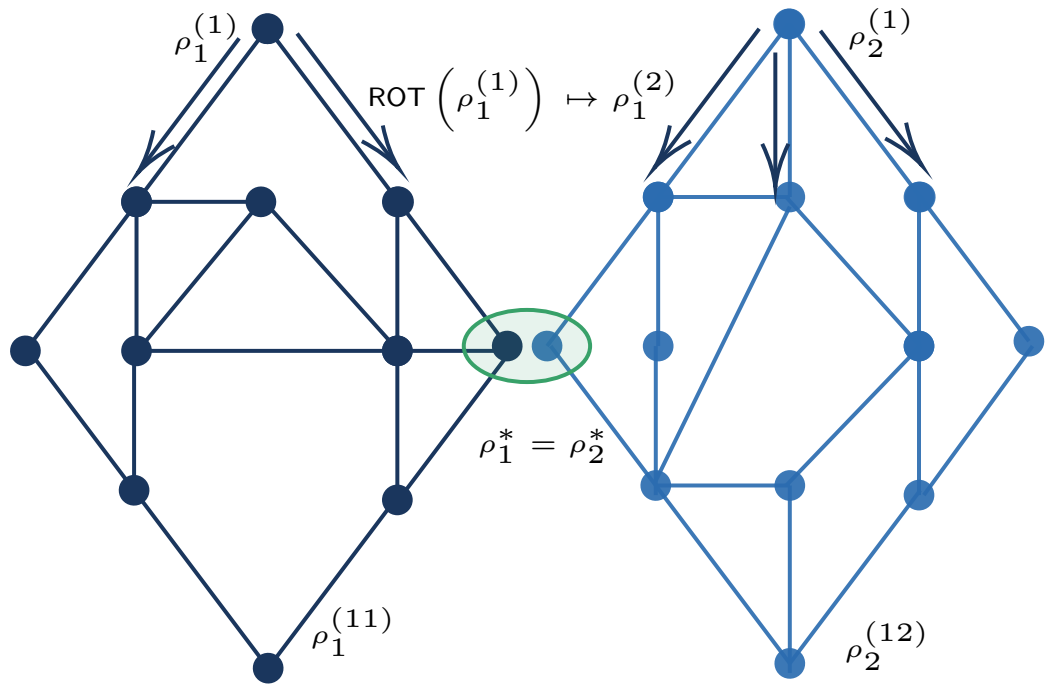
        \caption{\textbf{Lattice of Stable Matchings:} Two lattices share an
        intersection via matching $\matcH^*$. A lattice is constructed from
        each starting configuration $\matcH_i$ and $\matcH_j$, and
        via GS we can arrive at another stable matching.}
        \label{fig:gs_lattice}
    \end{subfigure}
    \caption{\textbf{Construction of Lattice Island Structure $\isL$:} Fundamentally, every lattice island on the lattice island structure $\isL$ is formed by intersection of a common stable matching within the lattices in the lattice set. }
    \label{fig:side_by_side}
\end{figure}

With respect to super-stability (Def.~\ref{def:super_stability}), some restricted preference models can allow for complex stable matching structures. More broadly, super-stability is closely related to lattice structure. Robust stable matchings, which remain stable across nearby instances, are known to form well-behaved structures under certain conditions, with sharp thresholds where this structure breaks down \citep{cheng:2023_stable,gangam:2026_robust_sm,mai:2018_finding_sm}.

\paragraph{Definitions and Notation:}

To combat the combinatorial complexity of exploring the full space of two-sided preference orderings, we introduce a computational budget $\rrBudget$ and a family of \emph{partition functions} $\{\partFunc : \parOrder \mapsto \parOrder'\}$ that refines a bipartite ordering topology, adding more elements to the new set $\parOrder'$. For a single agent $a$, we write $\partFunc_a : \succ_a \mapsto \succ_a'$ for the partition function acting on $a$'s ordering, and symmetrically $\partFunc_b$ for $b$. The integer $N_\partFunc$ denotes the number of partitions induced by $\partFunc$ on a given ordering. 

For each agent we let $\seT{P}^{(n)}$ denote the collection of partitions belonging to agent $n \in [N]$ on the A-side (and $\seT{P}^{(k)}$ for $k \in [K]$ on the B-side); the size of the $i$-th partition is $|\seT{P}^{(i)}|$. We write $\succ_a$ for the total ordering of an A-side agent $a$ over the $K$ B-side agents, and $\succ_b$ for the symmetric B-side ordering. The \emph{partial} ordering of $a$ over the B-side is denoted $\widehat{\succ}_a$, and similarly $\widehat{\succ}_b$; the hat denotes that only some pairwise relations have been established with high confidence, so the ordering is a refinement-in-progress of the true $\succ_a$. The completely uninformed ordering — a single block containing all opposing agents, with no relations resolved — is denoted $\breve{\succ}_i$; this is the starting state of any learning agent and corresponds to $C_\Psi = 0$ cuts.

\paragraph{Preference Configurations \& Partition Function $\partFunc$:}
% The partition function $\Psi_n(\cdot)$ is an abstract construct that maps any set to a collection of partial orderings, i.e. its \textit{partitions}. For example, referring to Fig. \ref{fig:partial_orderings_and_cuts}, in \(a_1\) there are three partitions of its ordering over \(b\). This is because \(b_1, b_2, b_3\) are mutually indifferent, while \(\{b_1, b_2, b_3\} \succ b_4 \succ b_5\) is strictly ordered. This yields three partitions, providing an abstract representation of partial orderings.
The partition function $\Psi_n(\cdot)$ is an abstract construct that maps a partial ordering to a partition of the opposing side into cells: agents within a cell are mutually indifferent, and the cells are what a strict complete refinement must order. For example, referring to Fig. \ref{fig:partial_orderings_and_cuts}, \(a_1\)'s partial ordering over the $b$'s has three cells, $\{b_1,b_2,b_3\}$, $\{b_4\}$, $\{b_5\}$, because \(b_1, b_2, b_3\) are mutually indifferent while \(\{b_1, b_2, b_3\} \succ b_4 \succ b_5\) is strictly ordered - this yields three cells. When these cells are ordered, the resulting ordered partition induces a total preorder on the opposing agents --- all agents in an earlier cell precede all agents in a later cell, and agents within a cell are tied. The partition is therefore the equivalence-class view of the partial ordering, not a partial order itself; a strict complete refinement of the partial order is a linear extension (a total order) on these cells.

A \emph{preference configuration} $\succ_{ab}^{(c)}$ for $(c) \in 1 \dots |\parOrder|$ is a fully-resolved two-sided ordering compatible with the current partial order $\parOrder$. Equivalently, the set of configurations is precisely the set of refinements of the partial order $\parOrder$. Each agent $n$ carries a set of $N_\partFunc$ partitions $\seT{P}^{(n)}$; the number of ways to fully order a single partition of size $|\seT{P}^{(n)}|$ is $|\seT{P}^{(n)}|!$, so the total number of configurations is,

\begin{align} \label{eq:pref_combinatorial_configs}
    |\parOrder| = \underbrace{\left(\prod^N_{n=1} \prod^{|\partFunc_n(B)|}_{i = 1} |\seT{P}^{(n)}_i(B)|! \right)}_{\text{A side}} \underbrace{\left(\prod^K_{k=1} \prod^{|\partFunc_k(A)|}_{j = 1} |\seT{P}^{(k)}_j(A)|! \right)}_{\text{B side}},
\end{align}

where $\partFunc_n(B)$ denotes the partition of the members on the $B$ side from the perspective of member $n$, i.e.,
$\partFunc_n(B):=\{\seT{P}_{1}^{(n)}(B),\ldots,\seT{P}_{n}^{(|\partFunc_n|)}(B)\},$ such that $|\bigcup_i \seT{P}^{(n)}_i(B)|=K$. Each partition consists of $n$ cells denoted as $\seT{P}_{i}^{(n)}$. Similarly, $\partFunc_k(A)$ denotes the partition of the members on the $A$ side from the perspective of member $k$.

\
\paragraph{Rotational Exploration and Lattice Construction:}
Applying Gale--Shapley to a fully-resolved configuration $\succ_{ab}^{(c)} \in \parOrder$ yields one stable matching $\matcH_m$; the map $\{\GS : \parOrder \to \SMSet, \; \parOrder^{(c)} \mapsto \matcH_m\}$ returns a stable matching from a fully-resolved preference, i.e., $\GS(\parOrder^{(c)})=\matcH_m$, where $\seT{M}$ denotes the set of matchings and $\SMSet \subseteq \seT{M}$ the stable subset. From any $\matcH_m \in \seT{M}$, remaining stable matchings compatible with the same configuration arise via rotations, producing a lattice $\laT^{(c)}$ among stable matchings. Each rotation is computable in polynomial time, and the number of unique rotations is upper-bounded by $\mathsf{nCr}(\bar{N}, 2)$ \citep{gusfield:1988_rotation}, where $\bar{N} \deF \max \{N, K\}$ (w.l.o.g. $N=K$ here for complexity analysis).  We denote each rotation as $\{\mathsf{ROT} : \matcH_m^{(i)} \to \matcH_m^{(j)} \}$, and a series of rotations generates a lattice. 
Thus a lattice $\laT^{(c)}$ is associated to each preference profile $\parOrder^{(c)}$. Importantly, two different preference profiles need not share the same rotation structure: even if their lattices share matchings, their rotations (and partial orders among rotations) can differ. We write $\setlaT$ for the set of all such lattices and denote the map $\mathsf{LAT} : \parOrder \to \setlaT$ sending any fully-resolved configuration to its corresponding lattice.

\paragraph{Intersection of Lattices and Islands $\mathsf{ConstructIslands}(\parOrder)$:}
Within $\setlaT$, distinct lattices may share vertices. Any matching in the intersection of two or more lattices is a \emph{centroidal lattice matching}, $\matcH^*$. When $V(\laT^{(i)}) \cap V(\laT^{(j)}) \neq \emptyset$, the lattices belong to the same \emph{island}, $\isL$. Precisely, $\isL^{(m)} \deF \{ \laT^{(i)}_m, \laT^{(j)}_m, \dots \}$ is a set of lattices connected by a single $\matcH^*_m$. A lattice may belong to multiple islands. We define a blocking function $\{\blockFunc: \setlaT \mapsto \isL \}$ mapping lattices to islands. For instance, $\isL = \blockFunc(\setlaT)$ might be $\{ \{\laT^{(i)}_m, \laT^{(j)}_m \}, \{\laT^{(i)}_n, \laT^{(j)}_n, \laT^{(k)}_n\}, \dots, \{\laT^{(i)}_o\} \}$. The blocking function partitions $\setlaT$ into islands, each sharing a common $\matcH^*$. In the worst case each lattice is its own island; in the best case all form one island. It involves searching all intersecting vertices and runs in polynomial time (Lem \ref{lem:intersection_islands_poly}; Algorithm \ref{alg:island_construction} in \S \ref{app:supp_alg}). The key advantage: given any $\parOrder$, if $|\isL| = 1$, the centroidal matching is a super stable matching (obviating resolution over all strict preferences). The number of intersecting matchings per lattice is also computable in polynomial time (Lemma \ref{lem:intersection_islands_poly} in \S \ref{app:intersection_islands_poly}). Thus $\isL$ is constructible, though worst-case configuration count scales exponentially with $|\parOrder| \le \bar{N}^2$, where $\bar{N} \deF \max\{N, K\}$. We denote the resulting procedure $\{\mathsf{ConstructIslands} : \parOrder \to \isL\}$.

\begin{theorem}\label{thm:super_stable_single_island}
\TXTSingleIslabsSuperStability
\end{theorem}

\begin{proof}
    For any $\isL^{(c)} \in \isL$, the centroidal lattice matching
    $\matcH^*_m$ is a stable matching for every refinement of every
    configuration $\parOrder \in \parOrder$ corresponding to a lattice in
    $\isL^{(c)}$. Therefore, if $|\isL| = 1$, every refinement of
    $\parOrder$ yields the same stable matching, which is \emph{super-stable} per Def. \ref{def:super_stability}.
\end{proof}

\subsection{Tractability of the Lattice Island Construction}

Eq. \eqref{eq:pref_combinatorial_configs}  shows that in the worst case a combinatorially large number of configurations leads to a combinatorially large number of candidate matchings. To control this growth, we impose a budget $\rrBudget$, and require that the number of configurations enumerated satisfies $|\setlaT| \le \rrBudget$. At each iteration, if the size of the candidate set of lattices is within the budget, we may enumerate all configurations explicitly. Otherwise, we subsequently break down the problem via partitioning until the configuration count falls below the budget (see \S \ref{sec:preflid}).

\vspace{-5pt}

\paragraph{Tractability via Optimal Partitioning:} Although enumerating all orderings in $\parOrder$ explodes factorially as demonstrated in Eq. \eqref{eq:pref_combinatorial_configs}, sufficient  partitioning can control the problem: generating tractable sub-problems can reduce complexity more than the factorial explosion, while incrementally resolving preference certainty among one agent at a time sequentially controls combinatorial explosion.  In the worst case, $\mathsf{ConstructIslands}(\parOrder)$ scales exponentially with $|\parOrder|$, therefore additional measures are needed to enforce tractability. In \S \ref{app:tractability_partition}, we demonstrate that complexity can be made tractable given a sufficient partition function (specifically Thm. \ref{thm:XPTractability}, \S \ref{app:tractability_partition}). As we introduce online learning in Sec. \ref{sec:online_learning}, the intuitive idea that sampling leads to partitions of $\parOrder$ and it is cheaper than offline combinatorial search, and the two complement each other (i.e. sample sufficiently to reduce the problem search space over $\isL$, and use the island structure to speed up sampling vice-versa).

\vspace{-7pt}

\section{Online Learning} \label{sec:online_learning}

\begin{figure*}[t]
\centering
\subfloat[Scenario: $N=K$.]{%
  \includegraphics[width=49mm]{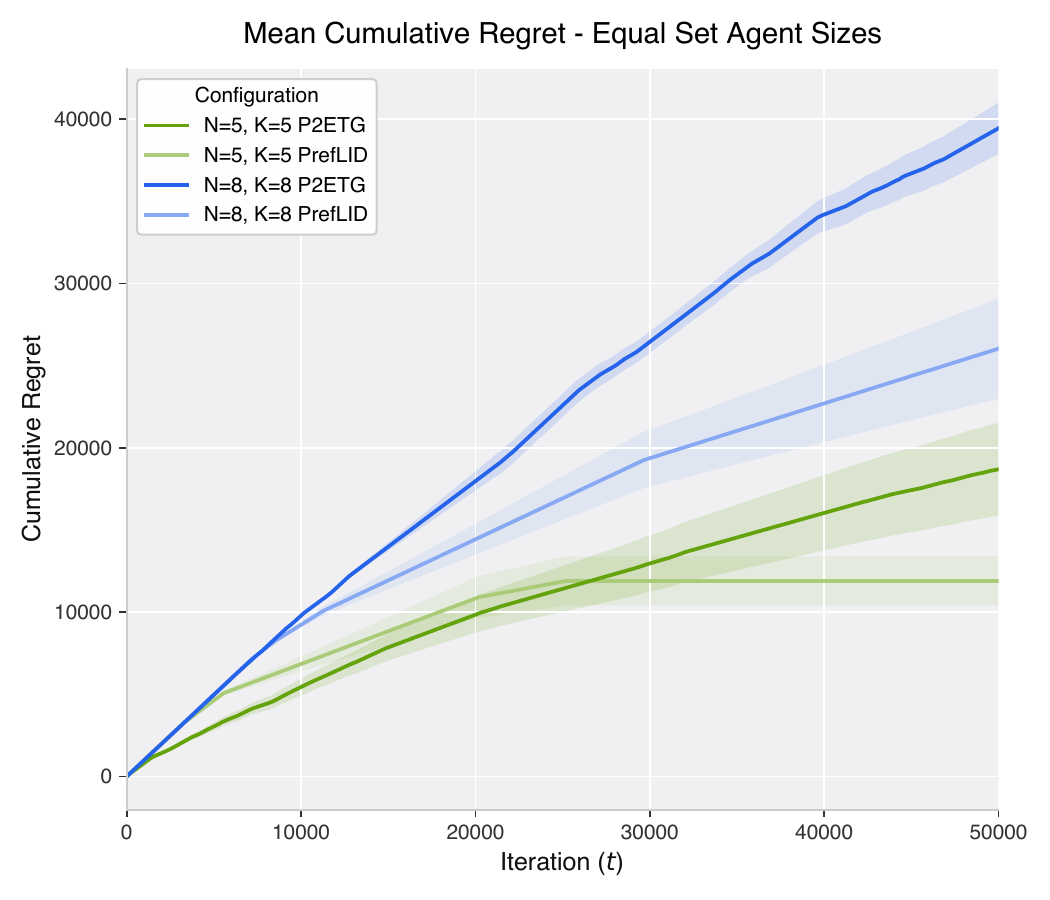}%
  \label{fig:reg_equal_nk}%
}\hspace{10pt}% Add gap here
\subfloat[Scenario: $N\neq K$.]{%
  \includegraphics[width=49mm]{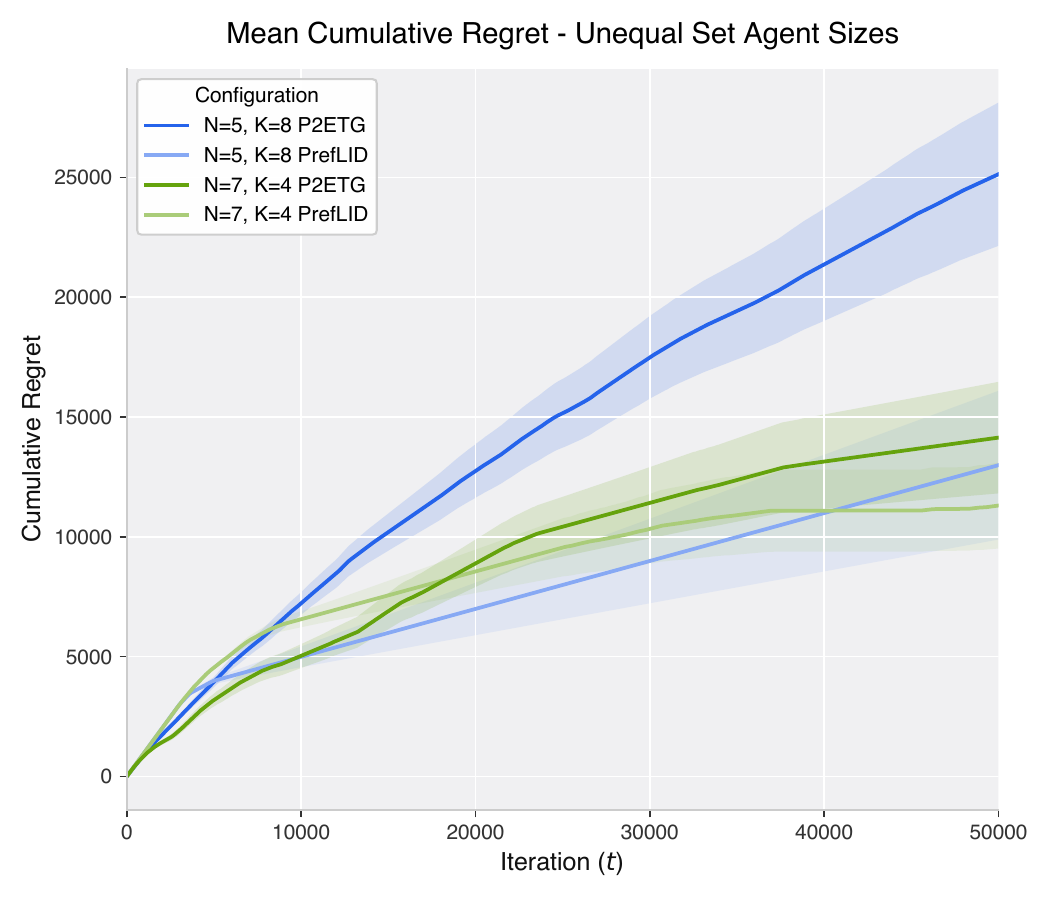}%
  \label{fig:reg_unequal_nk}%
}\hspace{10pt}% Add gap here
\subfloat[Scenario: LLM Routing.]{%
  \includegraphics[width=49mm]{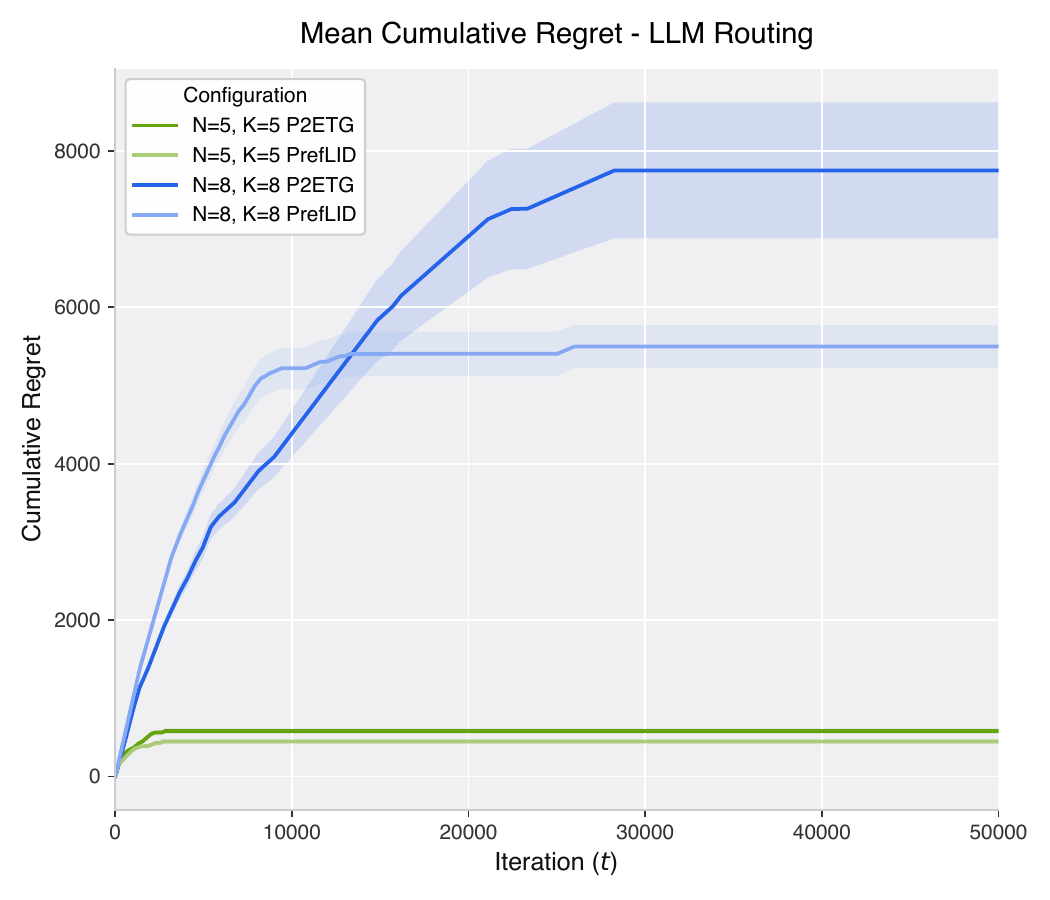}
  \label{fig:reg_llm}
}
\caption{
\textbf{Comparison of Cumulative Regret:} We evaluate $\mathsf{P2ETG}$ and $\mathsf{PrefLID}$ on both synthetically generated Bradley-Terry parameters, as well as LLM routing (see Sec. \ref{sec:llm_routing_main}). Across all instances, the cumulative regret performance of $\mathsf{PrefLID}$ versus $\mathsf{P2ETG}$ demonstrates lower cumulative regret for the same configuration of matching agents $(N, K)$. Shaded regions illustrate uncertainty up to one standard deviation. We present a more comprehensive ablation study in \S \ref{app:empirical}.
} \label{fig:all_regret_plots_main}
\vspace{-15pt}
\end{figure*}

\vspace{-5pt}

We extend optimisation under perfect information to a setting with parameter uncertainty, namely the parameters which govern agent preferences, $\parOrder$. Prior work typically uses an explore-then-commit (ETC) framework \citep{kong:23_many_to_one, kong:24_improved_analysis, liu:21_decentralized_matching}, drawing enough samples so that an estimate $\widehat{\succ}$ reflects the true preferences $\succ$ with high probability. 

Consequently, $\parOrder$ is not given explicitly but is
derived from the current confidence intervals: for each agent, we sort the
opposing agents by their estimated Bradley--Terry parameters
$\widehat{\theta}_{a,b}$ in descending order, and we place a \emph{cut}
between consecutive agents $b_i, b_{i+1}$ wherever their confidence
intervals are disjoint. Within a cell, the intervals overlap and the
internal order remains unresolved (refer to Figure \ref{fig:partial_orderings_and_cuts}).
Although the size of the space of configurations grows exponentially with the number of agents, we can compute the $|\parOrder|$, the cardinality of $\parOrder$, in polynomial time. 
\begin{lemma} \label{lem:size_of_BRT}
\TXTSizeBipartiteRankingTopology{(Proof in \S \ref{prf:size_of_BRT}.)}
\end{lemma}
\noindent \textit{Sketch of Proof:} The expression contains
$NN_\partFunc + KK_\partFunc = O(n^{2c})$ factorial factors. Since each
factor satisfies $|\seT{P}^{(i)}| \le M = O(n^c)$, every factorial can be
computed in $O(n^c)$ arithmetic operations. Thus all factorials require
$O(n^{3c})$ operations, while multiplying the resulting terms requires only
$O(n^{2c})$ additional operations. Hence $|\parOrder|$ is computable in
polynomial time in the input size $n$.

We present our initial baseline algorithm, $\mathsf{P2ETG}$ a key extension of an ETC bandit framework with additional two-sided uncertainty, and also adapts the Bradley--Terry pairwise comparison model directly into the learning protocol (see Sec. \ref{sec:p2etg}). We further devise $\mathsf{PrefLID}$, an improvement over $\mathsf{P2ETG}$, which reduces the sample complexity of ETC framework via construction of the island structure $\{\mathsf{ConstructIslands} : \parOrder \to \isL\}$ and targeted sampling (see Sec. \ref{sec:preflid}).

\vspace{-2pt}

\paragraph{Interaction Protocol:} We now extend from optimisation under perfect information to parameter uncertainty, where both sides may not know their own preferences and must learn them while obtaining a stable matching. We adopt an explore-then-commit framework, drawing enough samples so that an estimate $\widehat{\succ}_a$ reflects the true preferences $\succ_a$ with high probability, similar to the approach in \citep{kong:23_many_to_one, kong:24_improved_analysis, liu:21_decentralized_matching}. We adopt this framework with two extensions: we allow \emph{both} sides to be uncertain, and we embed the Bradley--Terry pairwise comparison model directly into the online learning algorithm and its regret analysis. Works such as \citep{kong:24_improved_analysis} hinge on observability of a utility function, but in matching a utility can typically only be compared via a measure over an outcome (e.g.\ from $a$'s perspective, matching $b_i$ yields 7.7 points in expectation versus 5.8 for $b_j$), which induces a ranking (e.g.\ $7.7 > 5.8 \implies b_i \succ b_j$). % Our framework obviates the need for a direct measure of the matching outcome. 

\paragraph{Feedback Mechanism:}  Building on \citep{kong:23_many_to_one, kong:24_improved_analysis, liu:21_decentralized_matching}, in our setting the bandit-feedback model is comparative rather than a real-valued reward. Instead of a sub-Gaussian reward $\mu_a : (a, b) \mapsto \mathbb{R}$ with a stationary mean $\expeC[\mu_a(a,b)]$ when $(a,b)$ is matched, a more natural approach governs the interaction. 

At each round, a set of pairwise comparisons are determined by the system - constituting a matching, $\matcH_t$. When a match between $(a, b)_t$ occurs, within $\matcH_t$,  the learner observes a binary comparison outcome $\bigl(o_{(a,b)},\, o_{(b,a)}\bigr)_t$, where a latent strict preference exists for both sides $\succ_{ab}$. This true ordering, $\succ_{ab}$, is unobserved, and is the goal of our learning algorithm. For example, the latent as the latent preference for $b_1 \succ_a b_2$ exists, then the probability that $\proB \bigl(o_{(a, b_1)} \succ_a o_{(a, b_2)} \bigr) > 0.5$. Notice we are comparing the outcomes of the match ups, where outcomes can be ranked - for example, For example, a firm interviewing candidate $b_1$ may receive a stronger assessment signal than from interviewing $b_2$, inducing $b_1 \succ_a b_2$. In each round robin round, each player on each side will have been matched with every member of the other set (for example, each member of $\seT{A}$ is matched with $\seT{B}$).

This preference over outcomes is governed by their internal preference parameters $\theta_a$ in accordance with the Bradley--Terry model (see \S \ref{sec:bradley_terry}), we denote this as $\proB_a(i \succ j)$. Evidently, each agent has an internal set of parameters $\theta_a$ which governs their preference function on each specific outcome from matching with other agents.

\paragraph{Definition of Regret:}

Our notion of regret applies the optimal benchmark of attaining \textbf{any} stable matching, expressed as,
\begin{equation} \label{eq:reg_def}
    \reG \deF \sum_{t=1}^T
    \indiC \left(\matcH_t \notin \SMSet \right).
\end{equation}
Here, regret declares a maximum loss of $1$ each time the algorithm fails to
obtain the \textbf{any} stable matching - which could be multiple thereof.
Previous work in this area typically defines regret with respect to the
\textbf{single} proposer-optimal, receiver-pessimal stable matching $\matcH_{GS}$, as
given by the Gale--Shapley algorithm \cite{kong:23_many_to_one, KongLi2024}. In our setting, we need not adhere to any specific stable matching, we measure regret simply as whether or not a stable matching has been obtained, i.e. no regret occurs when $\matcH_t \in \SMSet$.
% Misspecification can drive the
% algorithm to produce an alternative matching $\matcH'$, regardless of
% whether $\matcH'$ is stable.

\begin{definition}
    \textbf{Stopping Time:} The stopping time $\Tstop$ is the number
    of samples required such that, with probability at least $1 - \eta$,
    the learner can commit to a fixed strategy of play for all
    iterations $t > \Tstop$.
\end{definition}

% \begin{definition}
%     \textbf{Stopping Condition:} The stopping time $\Tstop$ is the number of samples required such that, with
%     probability at least $1 - \eta$, the learner can commit to a fixed
%     strategy of play for all iterations $t > \Tstop$.
% \end{definition}

\subsection{The Bradley--Terry Model Model for Preference Relation} \label{sec:bradley_terry}

The Bradley--Terry model is a probability model for
paired comparisons \citep{bradley1952rank}. It provides a principled framework for estimating the
relative strengths or abilities of items (players, teams, products, etc.) based
on observed pairwise contest outcomes. For a pair of items $i$ and $j$ with
positive strength parameters $\theta_i$ and $\theta_j$, the model specifies the
probability that $i$ beats $j$ in a contest as $\proB(i \succ j) = \theta_i/(\theta_i + \theta_j)$ \citep{Yao:1999_AsymptoticsWT}. It follows that $\theta_i > \theta_j$ implies $i \succ j$ in the preference
ranking over $\seT{B}$ for the corresponding agent. However, since
$\proB(i \succ j)$ is a probability, the closer $\theta_i$ and $\theta_j$ are to
each other, the more samples are required to distinguish the minute preference.
We refer to this as the \emph{preference parameter gap}, $\Delta$. In our two-sided setting, the parameters are indexed by both the agent and the partner, for example each A-side agent $a \in \seT{A}$ carries a vector
          $\theta_a \in \mathbb{R}_{>0}^{A}$, with coordinate
          $\theta_{a,b}$ describing $a$'s latent affinity for partner
          $b \in \seT{B}$, and mutatis mutandis for the B-side. Within agent $a$'s own ranking, the comparison between two partners $b_1, b_2 \in \seT{B}$ is then governed by,
          
\begin{equation} \label{eq:bt_local}
    \proB_a(b_1 \succ b_2)
    = \frac{\theta_{a,b_1}}{\theta_{a,b_1} + \theta_{a,b_2}} ,
\end{equation}

and symmetrically for the B-side\footnote{Often in literature, we can also express $\proB_a(b_1 \succ b_2) = (1 + \exp(\gamma_{a,b_1} \gamma_{a,b_1} + \gamma_{a,b_2}))^{-1}$, here by convention we use $\theta_{a,b} \deF \exp(\gamma_{a,b})$, aligning with the representation in  \cite{Yao:1999_AsymptoticsWT}.}. Because the two sides hold \emph{independent}
parameter vectors, the learning problem is genuinely two-sided: no side has
access to the ground truth. The standard method to estimate Bradley Terry parameters via maximum liklihood estimation is presented in \S \ref{sec:BT_estimation}.

\paragraph{Relation to Plackett--Luce.} Under Assumption~\ref{ass:reg_BT_params}, $\theta_a$ is a Plackett--Luce choice vector and Eq.~\eqref{eq:bt_local} is its two-element case. Relative (Plackett--Luce) feedback in matching-market bandits is due to \citet{KongEtAl2026}, whose subset rule reduces to Eq.~\eqref{eq:bt_local} at $|\mathcal N'|=2$.

\begin{assumption} \label{ass:reg_BT_params}
    \textbf{Bradley--Terry Parameter Normalisation.} The parameters of each agent are normalised so that
    $\sum_{b \in \seT{B}} \theta_{a,b} = 1$ for every $a \in \seT{A}$, and
    symmetrically $\sum_{a \in \seT{A}} \theta_{b,a} = 1$ for every
    $b \in \seT{B}$.
\end{assumption}

\subsection{Pairwise Two-Sided Explore-then-Gale-Shapley Algorithm}\label{sec:p2etg}

We now describe the \emph{Pairwise Two-Sided Explore-then-Gale--Shapley} algorithm ($\mathsf{P2ETG}$). Fundamentally, this is an adaptation of the explore-then-commit (ETC) algorithm adapted to pairwise comparisons via the Bradley Terry framework. The procedure runs in doubling epochs $\ell = 0, 1, 2, \dots$, where in epoch $\ell$ the target number of pairwise replications per agent--pair is $R_\ell = 2^\ell$. Within each epoch, all agents on both sides perform round-robin pairwise comparisons. After each epoch
the Bradley--Terry maximum-likelihood estimates are refreshed, confidence intervals are recomputed, the empirical preference profile is reconstructed by sorting $\hat{\theta}$ in descending order, and Gale--Shapley is run on the estimated profile. If the confidence intervals of every pair of potential partners have become disjoint for every agent, the algorithm commits to the
current Gale--Shapley matching and performs no further exploration. We present the explicit outline of this Alg. \ref{alg:p2etg} in \S \ref{app:p2etg}, and key empirical results in Fig. \ref{fig:all_regret_plots_main}.

\begin{theorem}\label{thm:PTwoETCGSRegret}
    \TXTPTwoETCGSRegret{(Proof is provided in \S \ref{prf:PTwoETCGSRegret}.)}
\end{theorem}

\subsection{Preference Resolution via Lattice Island Detection ($\mathsf{PrefLID}$)} \label{sec:preflid}

\begin{algorithm}[h!]
\caption{Abridged Version of Preference Resolution via Lattice Island Detection
($\mathsf{PrefLID}$) - See \S \ref{app:preflid}. }
\label{alg:candidate_pref_res}
\begin{algorithmic}[1]
% \Require Agent sets $\seT{A},\seT{B}$; budget $\rrBudget$; confidence
%          $\eta\in(0,1)$.
% \Ensure A certified matching $\matcH^*$.
\State Initialize the unresolved profile $\parOrder$ and all
       comparison statistics.
\Loop
    \If{the compatible lattice representation exceeds $\rrBudget$}
        \State Choose the next entering center agent $c$. % according to the $\mathsf{RRT}$ schedule.
        \State Perform $\mathsf{RRT}(c)$; update Bradley--Terry estimates,
               confidence intervals, and the blocks of $\parOrder$.
        \State \textbf{continue}
    \EndIf
    \State $\setlaT\gets\{\mathsf{LAT}(\succ_{ab}):
           \succ_{ab}\text{ is a strict refinement of }\parOrder\}$.
    \State $\isL\gets\mathsf{ConstructIslands}(\setlaT)$.
    \If{all surviving lattices share a certificate $\matcH^*$}
        \State \Return $\matcH^*$.
    \EndIf
    \State Select an unresolved comparison associated with competing islands
           (using the widest-confidence rule).
    \State Sample it until the relation is resolved with confidence $1-\eta$;
           update $\parOrder$.
\EndLoop
\end{algorithmic}
\end{algorithm}

$\mathsf{PrefLID}$ is a two-sided, explore-then-commit learner that certifies \emph{global} stability of a committed matching by verifying that all configurations compatible with the current uncertainty lead to a single lattice island. It improves on the sample complexity of $\mathsf{P2ETG}$ by certifying a matching at the level of refinements rather than at the level of individual pairs. The stopping condition triggers when $|\isL| = 1$, i.e., when only one island exists in the blocks of $\setlaT$, denoted as $\isL = \blockLat$; this implies, with probability $1 - \eta$, that the candidate matchings $\{ \matcH^{*}_m, \matcH^{*}_n, \dots \matcH^{*}_l \}$ corresponding to that singleton island are the true stable matchings. In other words, all rotations starting from the various possible estimated preference configurations share a common destination matching. A detailed exposition of $\mathsf{PrefLID}$ is provided in \S \ref{app:preflid} along with full algorithmic details. Key empirical results are presented in Fig. \ref{fig:all_regret_plots_main}.

\paragraph{Targeted Round-Robin Sampling (RRT):} $\mathsf{PrefLID}$ employs \emph{Targeted Round-Robin (RRT) Sampling} (see \S \ref{alg:rrt_round}) rather than sampling every pair uniformly. Specifically, each round centers on a single agent $c$, who performs a full round-robin against the opposite side ($K(K-1)/2$ comparisons per round), refining only $c$'s partition structure.

\begin{theorem}
\label{thm:preflid_dominance}
\textbf{Regret and Stopping Dominance of $\mathsf{PrefLID}$:}
Under the same preference--feedback model and confidence sequences as $\mathsf{P2ETG}$,
\[
  \reG^{\mathsf{PrefLID}}(T) = O\!\left(\reG^{\mathsf{P2ETG}}(T)\right),
  \qquad
  \Tstop^{\mathsf{PrefLID}} \le \Tstop^{\mathsf{P2ETG}},
  \qquad
  \Delta^{\mathsf{PrefLID}} \le \Delta^{\mathsf{P2ETG}},
\]
where $\Delta^{\mathsf{PrefLID}}, \Delta^{\mathsf{P2ETG}}$ are the smallest preference gaps
that must be resolved before the respective stopping conditions are certified.
\end{theorem}

\begin{proof}
$\mathsf{PrefLID}$ shares the sampling rule of $\mathsf{P2ETG}$, but resolves only comparisons that can distinguish between islands. Pairs whose ordering is irrelevant to the surviving centroidal matching may remain unresolved indefinitely without affecting the certified output. Hence the set of gaps $\mathsf{PrefLID}$ must resolve is a subset of those required by $\mathsf{P2ETG}$, giving $\Tstop^{\mathsf{PrefLID}} \le \Tstop^{\mathsf{P2ETG}}$ and $\Delta^{\mathsf{PrefLID}} \le \Delta^{\mathsf{P2ETG}}$. Moreover, a single island ($|\isL| = |\blockFunc(\setlaT)| = 1$) certifies by Theorem~\ref{thm:super_stable_single_island} that the shared centroidal matching $\matcH^*_1$ is super-stable, so $\mathsf{PrefLID}$ stops as soon as this holds. Since per-sample regret is identical under the shared sampling rule and $\mathsf{PrefLID}$ stops no later, $\reG^{\mathsf{PrefLID}}(T) \le C\,\reG^{\mathsf{P2ETG}}(T)$ for a universal constant $C$, yielding the stated regret order.
\end{proof}

\subsection{LLM Routing Markets - A Modern Application of Online Stable Matching}\label{sec:llm_routing_main}

\begin{figure}[t]
    \centering
    \includegraphics[width=0.85\linewidth]{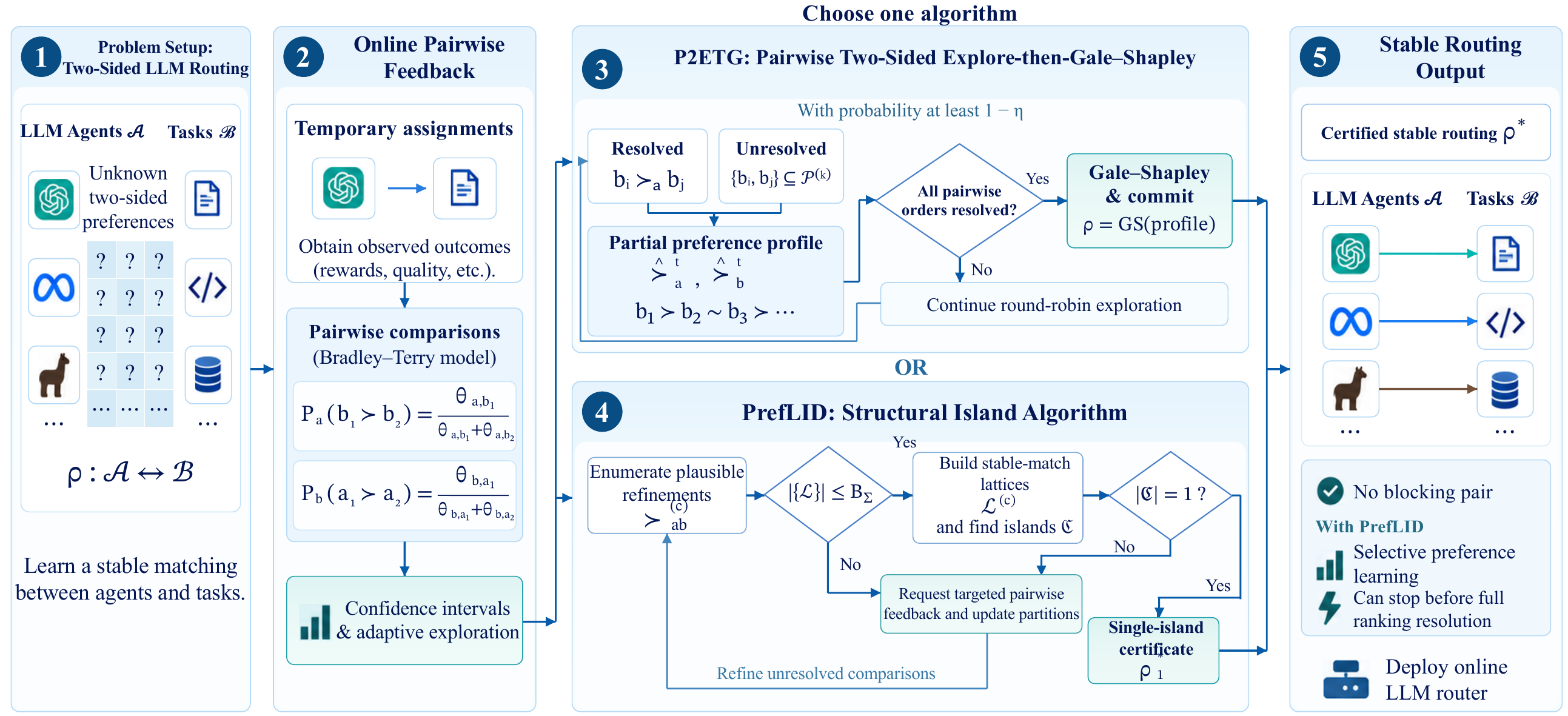}
    \caption{\textbf{The LLM Routing Workflow:} Both tasks and LLM agents have reciprocal preferences about which model is most appropriate for a given task. The LLM router’s goal is to match the optimal agent to each incoming task. Both $\mathsf{P2ETG}$ and $\mathsf{PrefLID}$ are benchmarked within this experimental setting. }
    \label{fig:llm_workflow}
    % \vspace{-20pt}
\end{figure}

We provide a novel application for the stable matching framework within the context of LLM routing, specifically instantiated from the \texttt{LLMRouterBench} evaluation suite~\citep{li2026llmrouterbench}. Here $N$ models (side A) and $K$ tasks (side B) rank each other by a shared compatibility, making their preferences reciprocal---$\text{rank}_a(b) = \text{rank}_b(a)$ for all $a, b$---since competence is mutual, simplifying the preference structure. This reciprocity collapses the ambiguity into a single unique stable matching, which the Gale--Shapley algorithm solves deterministically, returning the same unique outcome regardless of which side proposes. Computationally, the problem thus reduces to finding a singleton $\SMSet$. Therefore, the goal is to find $\{\rho^*\} = \SMSet$ in which no unpaired model--task pair prefers each other to their current partners (whereas arbitrary preferences can yield a rich lattice of stable matchings and task-optimal extremes). A workflow of this experimental setting is presented in Fig. \ref{fig:llm_workflow}. Key empirical results are presented in Fig. \ref{fig:all_regret_plots_main}, along with an extensive manifest of additional experiments and specifications can be found in \S \ref{sec:llm_experiments}.

\section{Conclusion}

We developed an online learning framework for two-sided stable matching via pairwise comparison. By representing uncertainty explicitly as a partial ordering --- whose indifference classes form a partition into cells, with refinements being total orders on those cells --- and exploiting the lattice--island structure of the induced stable matching space, we showed that a stable outcome can be certified without resolving every agent's complete ranking: our algorithm $\mathsf{PrefLID}$ resolves only those comparisons that distinguish between competing islands and thereby dominates the full-resolution baseline $\mathsf{P2ETG}$ in both stopping time and the smallest parameter gap that must be overcome, directly improving on cumulative performance. Our framework is particularly natural for modern AI systems that route queries among large language models, where preferences are learned from noisy pairwise or contextual feedback and fully ranking every model for every query type is prohibitively costly. Several directions remain open, such as, extending our framework to transient preferences with cycles, where pairwise comparisons may induce intransitive relations, would require rethinking both the partial-ordering representation and the notion of stability itself. Also, a finer parameterisation of regret via graph-structural properties of the island---such as its diameter, and connectivity---could yield sharper instance-dependent bounds and guide the design of even more sample-efficient exploration schedules.

\section*{AI Use Statement}

Generative AI tools were used to assist with manuscript editing, literature and citation checking, proof checking, and experimental design. AI-generated suggestions were independently reviewed and verified by the authors. All mathematical claims, proofs, algorithms, experiments, citations, and conclusions are the responsibility of the authors.

% \printbibliography

% \clearpage

\bibliography{main}                     % Your .bib file, no extension

@inproceedings{liu:21_decentralized_matching,
  title={Bandit learning in decentralized matching markets},
  author={Liu, Lydia T. and Ruan, Feng and Mania, Horia and Jordan, Michael I.},
  booktitle={International Conference on Learning Representations},
  year={2021},
  url={https://openreview.net/forum?id=LPJ_akD-cB}
}

@inproceedings{kong:24_improved_analysis,
  title={Improved analysis for bandit learning in matching markets},
  author={Kong, Fang and Wang, Zilong and Li, Shuai},
  booktitle={Advances in Neural Information Processing Systems},
  year={2024},
  url={https://proceedings.neurips.cc/paper_files/paper/2024/hash/a730abbcd6cf4a371ca9545db5922442-Abstract-Conference.html}
}

@article{athanasopoulos:25_learning_equilibria,
  title={Learning equilibria in matching markets from bandit feedback},
  author={Athanasopoulos, Andreas and Dimitrakakis, Christos},
  journal={arXiv preprint},
  volume={arXiv:2506.03802},
  year={2025},
  url={https://arxiv.org/abs/2506.03802}
}

@inproceedings{kong:23_many_to_one,
  title={Bandit learning in many-to-one matching markets},
  author={Kong, Fangyuan and Li, Shuai},
  booktitle={International Conference on Machine Learning},
  year={2023},
  url={https://proceedings.mlr.press/v202/kong23a.html}
}

@article{mirfakhar:26_relative_feedback,
  title={Bandit learning in matching markets with interviews},
  author={Mirfakhar, Amirmahdi and Wang, Xuchuang and Xu, Mengfan and Beyhaghi, Hedyeh and Hajiesmaili, Mohammad},
  journal={arXiv preprint},
  volume={arXiv:2602.12224},
  year={2026},
  url={https://arxiv.org/abs/2602.12224}
}

@inproceedings{liu2020competing,
  title     = {Competing Bandits in Matching Markets},
  author    = {Liu, Lydia T. and Mania, Horia and Jordan, Michael I.},
  booktitle = {Proceedings of the Twenty Third International Conference on Artificial Intelligence and Statistics},
  series    = {Proceedings of Machine Learning Research},
  volume    = {108},
  pages     = {1618--1628},
  year      = {2020},
  publisher = {PMLR}
}

@inproceedings{hosseini2024putting,
  title     = {Putting {G}ale \& {S}hapley to work: Guaranteeing stability through learning},
  author    = {Hosseini, Hadi and Roy, Sanjukta and Zhang, Duohan},
  booktitle = {Advances in Neural Information Processing Systems},
  volume    = {37},
  year      = {2024}
}

@article{gusfield:1988_rotation,
  title = {The structure of the stable roommate problem: Efficient representation and enumeration of all stable assignments},
  volume = {17},
  ISSN = {0097-5397},
  DOI = {10.1137/0217048},
  number = {4},
  journal = {SIAM Journal on Computing},
  publisher = {Society for Industrial and Applied Mathematics},
  author = {Gusfield, Dan},
  year = {1988},
  month = aug,
  pages = {742–769}
}

@article{Yao:1999_AsymptoticsWT,
title={Asymptotics when the number of parameters tends to infinity in the Bradley-Terry model for paired comparisons},
author={Simons, Gordon and Yao, Yi-Ching},
journal={The Annals of Statistics},
volume={27},
number={3},
pages={1041--1060},
year={1999},
publisher={Institute of Mathematical Statistics},
doi={10.1214/aos/1018031267},
url={https://projecteuclid.org/journals/annals-of-statistics/volume-27/issue-3/Asymptotics-when-the-number-of-parameters-tends-to-infinity-in-the/10.1214/aos/1018031267.full}
}

@article{bradley1952rank,
  title={Rank analysis of incomplete block designs: I. The method of paired comparisons},
  author={Bradley, Ralph Allan and Terry, Milton E},
  journal={Biometrika},
  volume={39},
  number={3/4},
  pages={324--345},
  year={1952},
  publisher={JSTOR}
}

@article{Zermelo:1929,
  author  = {Zermelo, Ernst},
  title   = {Die Berechnung der Turnier-Ergebnisse als ein Maximumproblem der Wahrscheinlichkeitsrechnung},
  journal = {Mathematische Zeitschrift},
  volume  = {29},
  pages   = {436--460},
  year    = {1929},
  doi     = {10.1007/BF01180541},
  url     = {https://doi.org/10.1007/BF01180541}
}

@article{Gale:1962,
  author  = {David Gale and Lloyd S. Shapley},
  title   = {College admissions and the stability of marriage},
  journal = {The American Mathematical Monthly},
  volume  = {69},
  number  = {1},
  pages   = {9--15},
  year    = {1962},
  doi     = {10.1080/00029890.1962.11989827}
}

@article{cheng:2023_stable,
  title={Stable matchings with restricted preferences: Structure and complexity},
  author={Cheng, Christine T and Rosenbaum, Will},
  journal={ACM Transactions on Economics and Computation},
  volume={10},
  number={3},
  pages={1--45},
  year={2023},
  publisher={ACM New York, NY}
}

@article{gangam:2026_robust_sm,
  title={Robust Stable Matchings: Dealing with Changes in Preferences},
  author={Gangam, Rohith Reddy and Mai, Tung and Raju, Nitya and Vazirani, Vijay V},
  journal={arXiv preprint arXiv:2601.07959},
  year={2026}
}

@article{mai:2018_finding_sm,
  title={Finding stable matchings that are robust to errors in the input},
  author={Mai, Tung and Vazirani, Vijay V},
  journal={arXiv preprint arXiv:1804.00553},
  year={2018}
}

@article{irving:1994_super_stable,
  author    = {Irving, Robert W.},
  title     = {Stable marriage and indifference},
  journal   = {Discrete Applied Mathematics},
  volume    = {48},
  number    = {3},
  pages     = {261--272},
  year      = {1994},
  publisher = {Elsevier},
  doi       = {10.1016/0166-218X(92)00179-P},
  mrclass   = {05C70},
  mrnumber  = {1261123},
  zbl       = {0796.05078}
}

@inproceedings{KongLi2023,
  author    = {Fang Kong and Shuai Li},
  title     = {Player-Optimal Stable Regret for Bandit Learning in Matching Markets},
  booktitle = {Proceedings of the 2023 Annual ACM-SIAM Symposium on Discrete Algorithms},
  pages     = {1512--1522},
  year      = {2023},
  doi       = {10.1137/1.9781611977554.ch55}
}

@inproceedings{LiWangKong2025,
  author    = {Shuai Li and Zilong Wang and Fang Kong},
  title     = {A Survey on Bandit Learning in Matching Markets},
  booktitle = {Proceedings of the Thirty-Fourth International Joint Conference on Artificial Intelligence},
  pages     = {10546--10554},
  year      = {2025},
  doi       = {10.24963/ijcai.2025/1171}
}

@article{KongEtAl2026,
  author  = {Fang Kong and Xiaoxi Zhang and Xiao Huang and Lijun Zhang and Shuai Li},
  title   = {Bandit Learning in Matching Markets with Relative Feedback},
  journal = {Theoretical Computer Science},
  volume  = {1071},
  pages   = {115846},
  year    = {2026},
  doi     = {10.1016/j.tcs.2026.115846}
}

@inproceedings{SuiEtAl2018,
  author    = {Yanan Sui and Masrour Zoghi and Katja Hofmann and Yisong Yue},
  title     = {Advancements in Dueling Bandits},
  booktitle = {Proceedings of the Twenty-Seventh International Joint Conference on Artificial Intelligence},
  pages     = {5502--5510},
  year      = {2018},
  doi       = {10.24963/ijcai.2018/776}
}

@inproceedings{BusaFeketeEtAl2014,
  author    = {R{\'o}bert Busa-Fekete and Bal{\'a}zs Sz{\"o}r{\'e}nyi and Eyke H{\"u}llermeier},
  title     = {{PAC} Rank Elicitation through Adaptive Sampling of Stochastic Pairwise Preferences},
  booktitle = {Proceedings of the Twenty-Eighth AAAI Conference on Artificial Intelligence},
  year      = {2014},
  doi       = {10.1609/aaai.v28i1.8978}
}

@inproceedings{ChenEtAl2020,
  author    = {Wei Chen and Yihan Du and Longbo Huang and Haoyu Zhao},
  title     = {Combinatorial Pure Exploration for Dueling Bandit},
  booktitle = {Proceedings of the 37th International Conference on Machine Learning},
  series    = {Proceedings of Machine Learning Research},
  volume    = {119},
  pages     = {1531--1541},
  year      = {2020}
}

@inproceedings{DrummondBoutilier2014,
  author    = {Joanna Drummond and Craig Boutilier},
  title     = {Preference Elicitation and Interview Minimization in Stable Matchings},
  booktitle = {Proceedings of the Twenty-Eighth AAAI Conference on Artificial Intelligence},
  pages     = {645--653},
  year      = {2014},
  doi       = {10.1609/aaai.v28i1.8829}
}

@article{AzizEtAl2020,
  author  = {Haris Aziz and P{\'e}ter Bir{\'o} and Serge Gaspers and Ronald de Haan and Nicholas Mattei and Baharak Rastegari},
  title   = {Stable Matching with Uncertain Linear Preferences},
  journal = {Algorithmica},
  volume  = {82},
  pages   = {1410--1433},
  year    = {2020},
  doi     = {10.1007/s00453-019-00650-0}
}

@article{AzizEtAl2022,
  author  = {Haris Aziz and P{\'e}ter Bir{\'o} and Tam{\'a}s Fleiner and Serge Gaspers and Ronald de Haan and Nicholas Mattei and Baharak Rastegari},
  title   = {Stable Matching with Uncertain Pairwise Preferences},
  journal = {Theoretical Computer Science},
  volume  = {909},
  pages   = {1--11},
  year    = {2022},
  doi     = {10.1016/j.tcs.2022.01.028}
}

@inproceedings{MaiVazirani2018,
  author    = {Tung Mai and Vijay V. Vazirani},
  title     = {Finding Stable Matchings That Are Robust to Errors in the Input},
  booktitle = {26th Annual European Symposium on Algorithms},
  series    = {Leibniz International Proceedings in Informatics},
  year      = {2018},
  doi       = {10.4230/LIPIcs.ESA.2018.60}
}

@article{GangamEtAl2026,
  author  = {Rohith Reddy Gangam and Tung Mai and Nitya Raju and Vijay V. Vazirani},
  title   = {Robust Stable Matchings: Dealing with Changes in Preferences},
  journal = {arXiv preprint arXiv:2601.07959},
  year    = {2026}
}

@article{hu2024routerbench,
  title={Router{B}ench: A benchmark for multi-{LLM} routing system},
  author={Hu, Qitian Jason and Bieker, Jacob and Li, Xiuyu and Jiang, Nan and Keigwin, Benjamin and Ranganath, Gaurav and Keutzer, Kurt and Upadhyay, Shriyash Kaustubh},
  journal={arXiv preprint arXiv:2403.12031},
  year={2024}
}

@article{li2026llmrouterbench,
  title   = {LLMRouterBench: A Massive Benchmark and Unified Framework for LLM Routing},
  author  = {Li, Hao and Zhang, Yiqun and Guo, Zhaoyan and Wang, Chenxu and Tang, Shengji and Zhang, Qiaosheng and Chen, Yang and Qi, Biqing and Ye, Peng and Bai, Lei and Wang, Zhen and Hu, Shuyue},
  journal = {arXiv preprint arXiv:2601.07206},
  year    = {2026}
}

@article{liu:2026_online_moe,
  title={Online mixture of experts: No-regret learning for optimal collective decision-making},
  author={Liu, Larkin and Etesami, Jalal},
  journal={Advances in Neural Information Processing Systems},
  volume={38},
  pages={166827--166873},
  year={2026}
}

@inproceedings{
zhang2024decentralized,
title={Decentralized Two-Sided Bandit Learning in Matching Market},
author={YiRui Zhang and Zhixuan Fang},
booktitle={The 40th Conference on Uncertainty in Artificial Intelligence},
year={2024},
url={https://openreview.net/forum?id=RUazoBVXIN}
}

@inproceedings{KongLi2024,
  title={Improved bandits in many-to-one matching markets with incentive compatibility},
  author={Kong, Fang and Li, Shuai},
  booktitle={Proceedings of the Thirty-Eighth AAAI Conference on Artificial Intelligence},
  volume={38},
  number={12},
  pages={13256--13264},
  year={2024}
}
\bibliographystyle{plainnat}            % Standard natbib-compatible preprint style

\clearpage
\appendix

\section{Algorithmic Theory}

\subsection{Size of the Bipartite Ranking Topology} \label{app:size_of_BRT}

\textit{\TXTSizeBipartiteRankingTopology}

\subsection{Proof of Lemma \ref{lem:size_of_BRT}} \label{prf:size_of_BRT}

\begin{proof}
    The number of unique configurations (refinements) over $\parOrder$, outlined in Eq. \eqref{eq:pref_combinatorial_configs} is presented again,
    
    \begin{align} 
        |\parOrder| = \underbrace{\left(\prod^N_{n=1} \prod^{N_\partFunc}_{i = 1} |\seT{P}^{(i)}|! \right)}_{\text{A side}} \underbrace{\left(\prod^K_{k=1} \prod^{K_\partFunc}_{j = 1} |\seT{P}^{(j)}|! \right)}_{\text{B side}},
    \end{align}
    
    Let the input size be $n$. We assume that the input sizes $N,K \in O(n^c)$ correspondingly since $N_{\partFunc} \leq N$ and  $K_{\partFunc} \leq K$ implies $N_{\partFunc}, K_{\partFunc} \in O(n^c)$, for some constant $c \in \mathbb{Z}^+$. The total number of factors is,
    \[
    T = N \cdot N_{\partFunc} + K \cdot K_{\partFunc} \in O(n^{2c}).
    \]
    
    For each factor $x = |\seT{P}^{(i)}| \le M$, computing $x!$ requires $O(M)$ multiplications of integers of size at most $M!$. Since $M \in O(n^c)$, this is $O(n^c)$ operations per factorial. There are at most $T$ such factorials to compute, giving,
    \[
    O(T \cdot M) \subseteq O(n^{3c})
    \]
    multiplications total. Multiplying the $T$ resulting factorial values together requires $T-1$ additional multiplications, which is $O(n^{2c})$. Thus the total number of arithmetic operations is.
    \[
    O(n^{3c}),
    \]
    which is polynomial in $n$. Therefore, the entire expression for $|\parOrder|$ is computable in polynomial time. 
\end{proof}

% \subsection{Proof of Theorem \ref{thm:super_stable_single_island}} \label{prf:super_stable_single_island}

% \TXTSingleIslabsSuperStability

% \begin{proof}
%     \todo{Write this proof.} Simple in any case of uncertainty there is one stable matching that fits all the uncertainty
% \end{proof}

\subsection{Tractability \& Optimal Partitions} \label{app:tractability_partition}

\paragraph{Optimal Partitioning (Cuts):} Let $ C_\Psi $ be a constraint on the maximum number of allowable cuts which divide any uninformative ordering $\breve{\succ}_a, \forall a \in [N]$ to a specific new configuration for a specific agent, $\succ_a$. Producing a cut typically involves resolving uncertainty among agents such that $b_i \succ b_j$ for any agents belonging to disjoint partitions $b_i \in \seT{P}^{(i)}$ and $b_j \in \seT{P}^{(j)}$ (or vice versa), that is an ordering exists among partitions. Therefore the number of cuts is bounded $ 0 \leq C_\Psi \leq N-1 $. Assuming that computational effort needs to be expended to produce a cut, we would like to understand what is the optimal partition size $|\seT{P}^{(i)}|$ for any partition such that the total number of refinements over $\widehat{\succ}_a, \forall a \in [N]$ is minimised (mutatis mutandis for the B-side). Thm. \ref{thm:budget_cut_tradeoff} argues that the optimal partition is the one where there is an equal (or near-equal) number of agents in each partition.

\begin{theorem}\label{thm:budget_cut_tradeoff}    
    \TXTBudgetCutTradeoff{(Proof provided in \S \ref{prf:budget_cut_tradeoff}.)}
\end{theorem}

% We can see from Eq. \eqref{eq:pref_combinatorial_configs} that there exist in the worst case a combinatorially large number of configurations leading to a combinatorially large number of candidate stable matchings $\matcH_m$. To combat the combinatorial explosion the size of $\setlaT$, in the worst case, we impose a budget $\rrBudget$ in order to explore the total number of possible configurations, $|\parOrder|$. The allowable budget must be greater than some polynomial function with resepect to the size of $\mathsf{Poly}(|\setlaT|) \leq \rrBudget$. 

% Thus for each sequential breakage event, there is a possibility of the size of candidate matchings corresponding to the set of lattices $\setlaT_{T}$ is within the computational budget, $\setlaT_{T} \leq \rrBudget$. \lar{Our certificate for the Fixed-Parameter Tractable (FPT) algorithm for XP (Slicewise Polynomial).} 

\noindent \textit{Sketch of Proof:} The proof reduces the problem to minimizing over the allowable number of subsets $k \leq C_\Psi + 1$, using a known formula for the minimal cost of balanced partitions. It then establishes that the minimal cost $\Sigma_{\min}(k)$ is non-increasing in $k$ by showing that splitting a subset of size $m \geq 2$ into two nonempty parts does not increase the total cost, due to the convexity of the factorial function. Consequently, the minimum over all admissible $k$ is attained at the maximum value $k = C_\Psi + 1$, meaning the optimal strategy uses exactly $C_\Psi$ cuts. Substituting this value into the balanced partition formula yields the explicit closed-form expression for $B_{\min}(C_\Psi)$ in terms of floor and ceiling factorials.

\begin{theorem} \label{thm:complexity_set_lattice}
    \TXTComplexitySetLattice{(Proof provided in \S \ref{prf:complexity_set_lattice}.)}
\end{theorem}

\noindent \textit{Sketch of Proof:} The proof assumes such $C_\Psi(N)$ exists and derives $\log(C_\Psi m!) = O(N^{-1}\log N)$ for $m = N/(C_\Psi+1)+1$. Stirling's bound gives $m\log m = O(N^{-1}\log N)$, but if $C_\Psi \le N/2$, then $m\log m = \Omega(N/C_\Psi)$, forcing $C_\Psi = \Omega(N^2/\log N)$. If $C_\Psi$ is that large, then $\log(C_\Psi m!) \ge \log C_\Psi = \Omega(\log N)$, contradicting the earlier bound. Hence both cases are impossible, so no such $C_\Psi(N)$ exists.

Theorem \ref{thm:complexity_set_lattice} demonstrates that generating the lattice set $\setlaT$ from any bipartite ranking topology $\parOrder$ in the worst case, cannot admit a polynomial-time solution even when $C_\Psi$ is chosen optimally. This motivates the search over regimes in which the problem exhibits weaker forms of computational hardness, such as \emph{slicewise polynomial tractability}.

\begin{definition}\label{def:xp_tractability}
    \textbf{Slicewise Polynomial (XP) Tractability:} A parametrized problem is in the class of \emph{Slicewise Polynomial} ($\mathsf{XP}$) if there exist computable functions $f, g \colon \mathbb{N} \to \mathbb{N}$ and an algorithm $\mathsf{ALG}$, such that for every instance of input $(x, k)$, the algorithm determines the output in time
    \begin{align}
        \bigO\!\left( f(k) \cdot |x|^{g(k)} \right),
    \end{align}
    where $k$ is the parameter, $|x|$ is the input size, and both $f$ and $g$ depend only on $k$ and are independent of $|x|$.
\end{definition}

\begin{theorem}\label{thm:XPTractability}
    \TXTXPTractability{(Proof in \S \ref{prf:XPTractability})}
\end{theorem}

\noindent \textit{Sketch of Proof:} The partition function $\blockFunc$ actively selects the cut budget $C_\Psi = N$ (the maximal feasible value, since $C_\Psi \leq N-1$), which resolves all uncertainty by making every partition a singleton. With partition size $N/C_\Psi = 1$, each factorial collapses to $1! = 1$, and the inner problem becomes $\bigO(N^d \cdot 1) = \bigO(N)$, polynomial in $N$. The outer problem is similarly collapsed: since every agent has a strict ordering, there is exactly one configuration, so the outer exponent is bounded by a constant $C_d = 1$. Island construction remains polynomial via inverted indexing and union--find, running in $\bigO(N|L| \cdot \alpha(N))$. Combining these, $\mathsf{ConstructIslands}(\cdot)$ runs in polynomial time $\bigO(N^C)$ for some constant $C$. If the resulting decomposition yields a single island, the shared centroidal matching is super-stable. 

% Theorem \ref{thm:complexity_set_lattice} demonstrates the problem of generating the $\setlaT$ cannot admit a polynomial time solution even when selecting the optimal minimising value for $C_\Psi$. We look to the scenario where under some assumptions the problem can exhibit the problem of \textit{fixed parameter tractability.} 

% \begin{definition}
%     \textbf{Fixed Parameter Tractability (FPT):} A parametrized problem is in the class of Fixed Parameter Tractable ($\mathsf{FPT}$) if there exists a computable function $f: \mathbb{N} \to \mathbb{N}$, a constant $d \geq 0$, and an algorithm $\mathsf{ALG}$, such that for every instance of input $(x, d)$ an algorithm can determine the output in polynomial time, $\bigO\!\left(f(k) \cdot |x|^d\right)$.
% \end{definition}

\subsection{Proof of Theorem \ref{thm:budget_cut_tradeoff}} \label{prf:budget_cut_tradeoff}

\TXTBudgetCutTradeoff

\begin{proof}
We prove the theorem in several steps.
% \lar{Note $k= C_\Psi + 1$}

\paragraph{Reduction to Fixed Number of Subsets}

Let \( \mathcal{P}_k \) denote the set of all partitions of \( N \) elements into exactly \( k \) nonempty subsets. 
By Lemma \ref{lem:balanced_partition_mini}, for any \( k \in \{1, \dots, N\} \), the minimum cost over \( \mathcal{P}_k \) is
\[
\Sigma_{\min}(k) = r_k \left( \left\lceil \frac{N}{k} \right\rceil \right)! + (k - r_k) \left( \left\lfloor \frac{N}{k} \right\rfloor \right)!,
\]
where \( r_k = N \bmod k \).

Now, a constraint of at most \( C_\Psi \) cuts means we can use any number of subsets 
\( k \) such that:
\[
1 \leq k \leq C_\Psi + 1.
\]
Let \( \mathcal{K} = \{1, 2, \dots, C_\Psi + 1\} \) be the set of allowable subset counts. 
The minimum cost under this constraint is therefore:
\[
B_{\min}(C_\Psi) = \min_{k \in \mathcal{K}} \Sigma_{\min}(k).
\]

\paragraph{Monotonicity of \( \Sigma_{\min}(k) \)}

We claim that \( \Sigma_{\min}(k) \) is non-increasing in \( k \). That is, for any 
\( 1 \leq k < N \),
\[
\Sigma_{\min}(k+1) \leq \Sigma_{\min}(k).
\]

To prove this, take the balanced partition achieving \( \Sigma_{\min}(k) \) with sizes 
\( n_1, \dots, n_k \) where \( \max_i n_i - \min_i n_i \leq 1 \). Since \( k < N \), 
at least one subset has size \( m \geq 2 \). Pick any such subset and split it into 
two nonempty subsets of sizes \( a \) and \( m-a \) where \( 1 \leq a \leq m-1 \). 
This yields a partition with \( k+1 \) subsets whose cost is:
\[
\Sigma' = \Sigma_{\min}(k) - m! + a! + (m-a)!.
\]

By the convexity of the factorial function, for any \( m \geq 2 \) and \( 1 \leq a \leq m-1 \),
\[
a! + (m-a)! \leq m!,
\]
with equality only when \( m = 2 \) and \( a = 1 \). Therefore \( \Sigma' \leq \Sigma_{\min}(k) \). 
Since \( \Sigma_{\min}(k+1) \) is the minimum over all partitions with \( k+1 \) subsets, 
we have:
\[
\Sigma_{\min}(k+1) \leq \Sigma' \leq \Sigma_{\min}(k).
\]

Thus \( \Sigma_{\min} \) is non-increasing.

\paragraph{Optimality of Using All Available Cuts}

Since \( \Sigma_{\min}(k) \) is non-increasing in \( k \), the minimum over 
\( k \in \{1, \dots, C_\Psi + 1\} \) is achieved at the largest allowable value:
\[
\min_{1 \leq k \leq C_\Psi + 1} \Sigma_{\min}(k) = \Sigma_{\min}(C_\Psi + 1).
\]

Therefore, the optimal strategy is to use exactly \( C_\Psi \) cuts, producing 
\( k = C_\Psi + 1 \) subsets, and then balance these subsets as evenly as possible.

\paragraph{Derivation of the Explicit Formula}

Let \( k = C_\Psi + 1 \). By Lemma \ref{lem:balanced_partition_mini}, the minimal cost for exactly 
\( k \) subsets is:
\[
\Sigma_{\min}(k) = r (q+1)! + (k - r) q!,
\]
where
\[
q = \left\lfloor \frac{N}{k} \right\rfloor, \qquad r = N - qk = N \bmod k.
\]

Substituting \( k = C_\Psi + 1 \) gives:
\[
\boxed{
B_{\min}(C_\Psi) = 
r \left( \left\lceil \frac{N}{C_\Psi + 1} \right\rceil \right)! 
+ (C_\Psi + 1 - r) \left( \left\lfloor \frac{N}{C_\Psi + 1} \right\rfloor \right)!,
}
\]
with \( r = N \bmod (C_\Psi + 1) \).

\end{proof}

\subsection{Proof of Theorem \ref{thm:XPTractability}} \label{prf:XPTractability}

\TXTXPTractability

\begin{proof}
We establish that $\mathsf{ConstructIslands}(\cdot)$ admits a polynomial-time algorithm under a specific design choice of the cut budget $C_\Psi = N^d$, delivered by the partition function $\blockFunc$. The key insight is that $\blockFunc$ is not merely a passive refinement operator but an active design mechanism: it selects the cut budget, and by choosing $C_\Psi = N^d$, it collapses the inner problem to polynomial time and simultaneously caps the outer exponent at a constant. We assume throughout that $\blockFunc$ can be applied arbitrarily at cost logarithmic in the number of configurations $T$, proportional to $\Delta$; this is fair since the construction cost is exponential in $N$, while sampling via $f$ is cheap.

\paragraph{Design Choice of the Cut Budget:}
The partition function $\blockFunc$ selects the cut budget $C_\Psi$. This is a design choice, not an external parameter. We choose
\begin{align}
    C_\Psi = N^d
\end{align}
for some $d \in \mathbb{Z}^+$. By Theorem~\ref{thm:budget_cut_tradeoff}, the optimal budget under this choice is
\begin{align}
    B_{\min}(N^d) = r \left( \left\lceil \frac{N}{N^d + 1} \right\rceil \right)! + (N^d + 1 - r) \left( \left\lfloor \frac{N}{N^d + 1} \right\rfloor \right)!,
\end{align}
where $r = N \bmod (N^d + 1)$.

\paragraph{Collapse of the Inner Problem:}
For $d \geq 1$, the partition size satisfies
\begin{align}
    \left\lceil \frac{N}{N^d + 1} \right\rceil = \left\lceil N^{1-d} \right\rceil \leq C_d
\end{align}
for some constant $C_d$ depending only on $d$. Therefore each factorial term is bounded by $C_d!$, a constant independent of $N$, and the inner problem complexity satisfies
\begin{align}
    B_{\min}(N^d)^{4} \;\leq\; N^d \cdot C_d^{4} \;\in\; \bigO\!\left(N^d\right),
\end{align}
which is polynomial in $N$. The exponent is $4$, not $4N$: the factor of $N^2$ from the players and $N^2$ from the rotations contributes a constant-degree polynomial per configuration, not a per-agent exponential. The inner problem therefore grows polynomially in $N$, hence polynomially in the input size $n$.

\paragraph{Tightening $\blockFunc$ to Control the Outer Exponent:}
The outer problem---enumerating all configurations in $\parOrder$---would grow exponentially in $N$ if left unchecked. The partition function $\blockFunc$ is tightened as follows. Under the cheap partitioning assumption, $\blockFunc$ resolves at least one agent's preferences fully at each application, yielding $|\widehat{\succ}_c| = 1$ for some $c \in \setA \cup \setB$. Applying $\blockFunc$ repeatedly resolves all ties for that agent, and the cost of this operation is $\bigO(\log T) \leq \bigO(\log |\parOrder|)$, absorbed into the polynomial factor.

Crucially, the design choice $C_\Psi = N^d$ ensures that the number of partitions scales polynomially with $N$, while the partition size $N/C_\Psi = N^{1-d}$ remains bounded by the constant $C_d$. Consequently, the multiplicative growth factor introduced by each new agent is bounded by a constant that depends only on $C_d$, not on $N$. Iterating this procedure over all agents, the total exponential term is bounded by $C_d^{O(1)}$, a constant. The outer exponent is therefore effectively kept constant, and the outer problem contributes only a polynomial factor.

\paragraph{Sequential Agent Addition with Bounded Complexity:}
Consider a sequential method starting with $2$ agents and adding new entrants one at a time. Each entrant $c \in \setA \cup \setB$ introduces additional complexity $\parOrder \to \parOrder'$ as the number of possible refinements would ordinarily increase exponentially. However, at each addition, $\blockFunc(\parOrder)$ is applied to fully resolve at least one agent's preferences, collapsing that agent's contribution to a single strict ordering. Because the partition size is bounded by $C_d$, the multiplicative growth factor at each step is bounded by a constant depending only on $C_d$. Iterating this procedure, the total exponential term is bounded by a fixed constant $\tilde{N} = C_d$, independent of $N$. Therefore, as $N \to \infty$,

\begin{align}
    B_{\min}(N^d)^{4} \;\leq\; N^d \cdot C_d^{4} \;\in\; \bigO\!\left(N^d\right).
\end{align}

\paragraph{Island Construction is Polynomial:}
By Lemma~\ref{lem:intersection_islands_poly}, given the set of lattices $\setlaT$, the island decomposition $\isL = \blockFunc(\setlaT)$ can be computed in polynomial time via inverted indexing and union--find:
\begin{align}
    \bigO\!\left(N|L| \cdot \alpha(N)\right),
\end{align}
where $|L|$ is the number of vertices per lattice and $\alpha$ is the inverse Ackermann function.

\paragraph{Combining the Bounds:}
The total complexity of $\mathsf{ConstructIslands}(\cdot)$ is the product of the inner problem complexity (polynomial in $N$ under $C_\Psi = N^d$), the outer problem complexity (controlled by the tightened $\blockFunc$ with constant exponent $C_d$), and the island construction complexity (polynomial). Therefore,
\begin{align}
    \bigO\!\left(\mathsf{ConstructIslands}\right) \;\in\; \bigO\!\left(N^C\right)
\end{align}
for some constant $C$ depending on $d$ and $C_d$. This is polynomial in the input size $n$, and exponential scaling is eliminated entirely.

\paragraph{Super-Stability from a Single Island:}
If the algorithm yields a single island, i.e., $|\isL| = 1$, then by Theorem~\ref{thm:super_stable_single_island} all lattices in $\setlaT$ share a common centroidal lattice matching $\matcH^*_1$. This matching is stable for every refinement of $\parOrder$, which is the definition of super-stability.

Therefore, under the specific design choice $C_\Psi = N^d$ delivered by the partition function $\blockFunc$, and with $\blockFunc$ tightened to resolve at least one agent's preferences at each application, the algorithm $\mathsf{ConstructIslands}(\cdot)$ runs in polynomial time. The outer exponent is effectively kept constant by the design of $\blockFunc$, and exponential scaling is eliminated.

% \begin{align}
%     \bigO\!\left(\mathsf{ConstructIslands}\right) \;\in\; \bigO\!\left(N^C\right)
% \end{align}
%for some constant $C$ depending on $d$ and $C_d$. 

\end{proof}

\subsection{Proof of Theorem \ref{thm:complexity_set_lattice}} \label{prf:complexity_set_lattice}

\TXTComplexitySetLattice

\begin{proof}
    Assume such $C_\Psi(N)$ exists (per Theorem \ref{thm:budget_cut_tradeoff}). Then,
    \[
    \left(C_\Psi m!\right)^{2N} \le N^{O(1)}, \quad m=\frac{N}{C_\Psi+1}+1.
    \]
    
    Taking logarithms,
    
    \[
    2N\log(C_\Psi m!) = O(\log N).
    \]
    
    Thus $\log(C_\Psi m!) = O(N^{-1}\log N)$.  
    But always,
    
    \[
    \log(C_\Psi m!) \ge \log(m!) \ge c\, m\log m
    \]
    
    for some constant $c>0$ (Stirling). Hence
    
    \[
    m\log m = O\!\left(\frac{\log N}{N}\right).
    \]
    
    Now $m = \frac{N}{C_\Psi+1}+1$. If $C_\Psi \le N/2$, then $m \ge 2$, so
    
    \[
    m\log m \ge \frac{N}{C_\Psi+1}\log 2 = \Omega(N/C_\Psi),
    \]
    
    which cannot be $O(N^{-1}\log N)$ unless $C_\Psi = \Omega(N^2/\log N)$.  
    If instead $C_\Psi$ is that large, then $C_\Psi m! \ge C_\Psi$, so
    
    \[
    \log(C_\Psi m!) \ge \log C_\Psi = \Omega(\log N),
    \]
    
    contradicting $\log(C_\Psi m!) = O(N^{-1}\log N)$ for large $N$. Therefore no such $C_\Psi(N)$ exists. 
    
    \[
    \boxed{\nexists\, C_\Psi(N) : \left[C_\Psi\left(\frac{N}{C_\Psi+1}+1\right)!\right]^{2N} \le N^{O(1)}.}
    \]
\end{proof}

\subsection{Finding the Intersection(s) of Lattice Islands} \label{app:intersection_islands_poly}
\begin{lemma}\label{lem:intersection_islands_poly}
    \TXTPolyIslandIntersect{(Proof provided in \S \ref{prf:intersection_islands_poly}. )}
\end{lemma}

\noindent \textit{Sketch of Proof:} The proof is constructive. First,
build an inverted index $\isL(v) = \{i : v \in S_i\}$ in $O(N|L|)$ time.
A vertex $v$ is intersecting iff $|\isL(v)| \ge 2$. Then form a graph on
$\{S_1, \dots, S_N\}$, connecting all sets in $\isL(v)$ for each
intersecting $v$. Its connected components are exactly the islands.
Explicit edge addition costs $\sum_v k_v^2 \le N \cdot N|L| = O(N^2|L|)$;
using DSU to union all sets in each $\isL(v)$ reduces this to
$O(N|L| \cdot \alpha(N))$.

\subsection{Proof of Lemma \ref{lem:intersection_islands_poly}} \label{prf:intersection_islands_poly}

\TXTPolyIslandIntersect

\begin{proof}
We prove the lemma by giving a constructive algorithm with 
three polynomial-time steps.

\paragraph{Inverted Index \& Intersecting Vertices:}
Initialize a hash map \( \mathcal{I}: V \to 2^{[N]} \), where \( V \) is 
the universe of vertices. For each set \( S_i \) and each vertex 
\( v \in S_i \), append \( i \) to \( \mathcal{I}(v) \). 
This takes \( O(N|L|) \) time and space. A vertex \( v \) is intersecting iff \( |\mathcal{I}(v)| \ge 2 \). 
Collect all such vertices by scanning \( \mathcal{I} \). 
This also takes \( O(N|L|) \) time.

\paragraph{Island Construction:}
Build an undirected graph \( G \) whose vertices are the sets 
\( S_1, \dots, S_N \). For every intersecting vertex \( v \), 
add edges (or cliques) connecting all pairs of sets in 
\( \mathcal{I}(v) \). The connected components of \( G \) are 
the desired islands.

\paragraph{Complexity Analysis:}
Let \( k_v = |\mathcal{I}(v)| \). For each vertex \( v \), 
adding all pairwise edges costs \( O(k_v^2) \). 
Since \( \sum_v k_v = N|L| \) and each \( k_v \le N \), 
the total cost is at most:
\[
\sum_{v} k_v^2 \le \left( \max_v k_v \right) \cdot \sum_v k_v 
\le N \cdot N|L| = O(N^2 |L|),
\]
which is polynomial in the input size. Using a disjoint set union (DSU), instead of adding edges explicitly, we union all 
sets in \( \mathcal{I}(v) \) together. Each union costs 
\( O(\alpha(N)) \), so the total becomes:
\[
O(N|L| \cdot \alpha(N)),
\]
which is nearly linear.
\end{proof}

\clearpage

\section{Learning theory}

\subsection{Definition of Breakage Event}

\begin{definition}
    \textbf{Breakage Event:} Given a confidence parameter $\eta \in (0, 1)$, a \emph{breakage event} at round $t \in \{1, \dots, T\}$ between two agents $i$ and $j$, denoted as $\textup{\tT{Br}}(i, j)$, occurs when their confidence intervals no longer overlap. Formally, holding with probability at least $1 - \eta$, either the upper confidence bound of agent $j$ is strictly smaller than the lower confidence bound of agent $i$, or vice versa:
    \[
        \textup{\tT{Br}}(i, j): \quad \left( \text{UCB}_j^t < \text{LCB}_i^t \right) \quad \lor \quad \left( \text{UCB}_i^t < \text{LCB}_j^t \right)
    \]
    where $\text{UCB}_k^t = \hat{\theta}_k^t + r_k^t$ and $\text{LCB}_k^t = \hat{\theta}_k^t - r_k^t$ for $k \in \{i, j\}$ denote the confidence limits for the Bradley-Terry preference parameters $\theta_i$ and $\theta_j$, guaranteeing that the true parameters are statistically separated ($\theta_i > \theta_j$ or $\theta_j > \theta_i$) with high probability.
\end{definition}

\subsection{$\mathsf{P2ETG}$ Algorithm} \label{app:p2etg}

\begin{algorithm}[H]
\caption{Pairwise Two-Sided Explore-then-Gale-Shapley Algorithm ($\mathsf{P2ETG}$)}
\label{alg:p2etg}
\begin{algorithmic}[1]
\Require Agent sets $\mathcal{A}$ and $\mathcal{B}$ with $|\mathcal{A}|=N$ and $|\mathcal{B}|=K$; confidence level $\eta\in(0,1)$.
\State Initialize epoch index $\ell \gets 0$.
\State Initialize sample counts $n_{ij}\gets 0$ for all $(i,j)\in \mathcal{A}\times\mathcal{B}$.

\While{$t \leq T$}
    \State $R_\ell \gets 2^\ell.$ \Comment{Set the target number of round-robin replications:}
    % \State Run additional round-robin exploration until every pair $(i,j)\in\mathcal{A}\times\mathcal{B}$ has exactly $R_\ell$ observations.
    % \State $\text{For each }(i,j)\in\mathcal{A}\times\mathcal{B},\text{ sample }(i,j)
    % \text{ until }
    % n_{ij}^{(\ell)} = R_\ell .$
    \ForAll{$(i,j)\in\mathcal{A}\times\mathcal{B}$} \Comment{Round-Robin Exploration Phase}
        \While{$n_{ij}^{(\ell)} < R_\ell$}
            \State Sample pair $(i,j)$ and observe $X_{ij}^{(n_{ij}^{(\ell)}+1)}$.
            \State Update $n_{ij}^{(\ell)} \gets n_{ij}^{(\ell)}+1$, $t \gets t+1$
        \EndWhile
    \EndFor

    \State For every pair $(i,j)\in\mathcal{A}\times\mathcal{B}$, update the empirical estimate $\widehat{\theta}_{ij}$ \Comment{see \S \ref{sec:BT_estimation}.}
    
    % \[
    %     \widehat{\theta}_{ij}^{(\ell)}
    %     =
    %     \frac{1}{R_\ell}
    %     \sum_{r=1}^{R_\ell} X_{ij}^{(r)} .
    % \]

    \State Construct confidence intervals,
    $
    \isL_{ij}^{(\ell)}
        =
        \left[
        \mathrm{LCB}_{ij}^{(\ell)},
        \mathrm{UCB}_{ij}^{(\ell)}
        \right]
    $
        
    % according to the learner's chosen valid confidence rule.
    \State Construct the empirical preference profile $\widehat{\succ}^{(\ell)}$ from $\widehat{\theta}^{(\ell)}$.

    \State $\matcH_t \gets \mathrm{GS}\!\left(\widehat{\succ}^{(\ell)}\right)$ \Comment{Gale-Shapley Algorithm on estimated preferences.}
    
    \If{$\forall (i,j),\quad
        \mathrm{LCB}_{ij}^{(t)}>0 \ \lor\ \mathrm{UCB}_{ij}^{(t)}<0$} \Comment{The exploration stopping condition.}
        % \State That is, for every relevant comparison between $j$ and $j'$ for agent $i$, either
        % \[
        %     \mathrm{LCB}_{ij}^{(\ell)}
        %     >
        %     \mathrm{UCB}_{ij'}^{(\ell)}
        % \]
        % or
        % \[
        %     \mathrm{LCB}_{ij'}^{(\ell)}
        %     >
        %     \mathrm{UCB}_{ij}^{(\ell)} .
        % \]
        \State $T_{\mathrm{stop}}
            \gets
            NK R_\ell$.
        \State \textbf{break}.
    \Else {}
    \State Update epoch index $\ell\gets 
    \ell+1$.
    \EndIf
\EndWhile

\For{all future rounds $t>T_{\mathrm{stop}}$}
    \State Play the committed matching $\matcH_t$. \Comment{Perform no further round-robin exploration.}
\EndFor
\end{algorithmic}
\end{algorithm}

\subsection{Proof of Theorem \ref{thm:PTwoETCGSRegret}} \label{prf:PTwoETCGSRegret}

\TXTPTwoETCGSRegret

\begin{proof}
    With the sample complexity results, we derive the regret bound, via the ETC until $\Tstop$ argument. First, we define an event called \textit{breakage}, $\tT{Br}$ where,

    \begin{align}
        \forall (i,j),\quad \mathrm{LCB}_{ij}^{(t)}>0 \ \lor\ \mathrm{UCB}_{ij}^{(t)}<0.
    \end{align}

    This means that for all pairs, the estimated comparative advantage $\hat{\theta}_i - \hat{\theta}_j$ is with high confidence certain (no overlapping confidence bounds for any pair). Once breakage occurs, we are free to exit the exploration phase.
    
    Let us define a failure event as $F$. The regret can be decomposed into the following components,

    \begin{align}
        % \reG &= \expeC \left[ \sum_{t=1}^T \left( 1 - \indiC(\matcH_t \in \seT{M}) \right) \right] \\
        \reG &=\expeC \left[ \sum_{t=1}^T \indiC(\matcH_t \notin \seT{M}) \right] \\
        &\leq \sum_{t=1}^T  \proB(\matcH_t \notin \seT{M} | \neg F)  + \proB(F)T  \\
        &= \underbrace{NK \sum_{\ell=1}^{\log ( T_{\tT{stop}}) } 2^\ell}_{\text{Exploration Phase}} 
        + \underbrace{\sum_{t=\Tstop}^T \proB(t \geq \Tstop|\neg F \land \neg \tT{Br})}_{\text{Insufficient Sample Size}}  
        + \underbrace{\eta T}_{\text{ Failure}} \\
        &= NK \sum_{\ell=1}^{\log ( \Tstop) } 2^\ell + \sum_{t = \Tstop}^T \eta + \eta T
    \end{align}

    \paragraph{Regret Decomposition:} At each time step the maximum instantaneous regret is $1$. The regret is decomposed into 3 components, first the \textit{exploration phase} consists of round-robin exploration of all possible pair matching, and doubling the number of exploratory trials so long as breakage is not achieved. The second contribution to regret the risk of \textit{insufficient sample size}, after drawing more the $\Tstop$ samples. Here breakage has not occurred, and round robin exploration continues - in the worst case until $T$ rounds. Lastly, $\eta T$ represents the risk of \textit{catastrophic failure}, such that breakage $\tT{Br}$ has occurred, therefore obtaining high confidence estimates, but we've estimated them incorrectly, inducing catastrophic error - occurring with probability $\eta$. Selecting $\eta = 1/T$, we have,

    % \begin{align}
    %     \reG &\leq \sum_{\ell=1}^{\log ( \Tstop) } 2^\ell NK + \sum_{t=\Tstop}^T  \proB(\matcH_t \notin \seT{M} | \neg F \land \tT{noBreak} )  + 1 \\ 
    %     &\leq C + \sum_{t=1}^T \frac{NK}{T}
    % \end{align}

    \begin{align}
        \reG &\leq NK\sum_{\ell=1}^{\log ( \Tstop) }  2^\ell + \sum_{t = 1}^T \frac{1}{t} + 1 \\
        &= NK (2^{\log(\Tstop + 1)} - 2) + \log T + 1 \\
        &= NK(\Tstop + 1) - 2NK + \log T + 1 \\
        &= NK\Tstop + \log T + 1 - NK
    \end{align}

    Note $\sum_{t = \Tstop}^T \frac{1}{T} \leq \sum_{t = \Tstop}^T \frac{1}{t} \leq \sum_{t = 1}^T \frac{1}{t}$. The core of the proof is to show that the constant $T_{\tT{stop}}$ is not a variable of $T$. This is a consequence of Theorem \ref{thm:gs_sample_complexity}, where the for any parameter $\eta$, the sample complexity needed to obtain a confidence that catastrophic failure $\proB(F) \leq 1- \eta$ is,

    \begin{align} \label{eq:no_regret_sample_complex_T}
    T \geq \frac{
        128 M_K^2
        \log\left(
        2(K+1)T
        \right)}{K\Delta_{ij}^2}
    \end{align}

    selecting $\eta = 1/T$. Importantly, we demonstrate that the solution to Eq. \eqref{eq:no_regret_sample_complex_T}, denoted as $\Tstop$ exists as a finite real positive number. That is invariant of $T$ itself. Such that $\Tstop$ until the smallest gap can be distinguished between pairs $i$ and $j$.  
    
    \[
    T
    \geq
    \frac{256M_K^2}{K\Delta_{ij}^2}
    \log\!\left(
    \frac{256(K+1)M_K^2}{K\Delta_{ij}^2}
    \right) = \frac{C_1}{K\Delta_{ij}^2} \log \left( \frac{C_2}{K\Delta_{ij}^2} \right)
    \]

    Therefore, assuming sufficiently large sample size $T > \Tstop$ we can establish the regret bound,

    \begin{align}
        \reG \leq NK\frac{C_1}{K\Delta_{ij}^2} \log \left( \frac{C_2}{K\Delta_{ij}^2} \right) + \log(T) + 1 - NK
    \end{align}

To combine the above additive terms into a single term, let $C_1, C_2 > 0$ be absolute constants. Then, for all $N, K, T \in \mathbb{N}$ and $\Delta > 0$ such that $\frac{N C_1}{\Delta^2} \geq 1$,
\[
N \frac{C_1}{\Delta^2} \log \left( \frac{C_2}{K \Delta^2} \right)  + \log T 
\ \leq \
N \frac{C_1}{\Delta^2} \log \left( \frac{C_2 T}{K \Delta^2} \right).
\]

Since $\frac{N C_1}{\Delta^2} \geq 1$, it holds that
\[
\log T \ \leq \ \frac{N C_1}{\Delta^2} \log T.
\]
Therefore,
\begin{align*}
N \frac{C_1}{\Delta^2} \log \left( \frac{C_2}{K \Delta^2} \right) + \log T 
&\ \leq \ 
N \frac{C_1}{\Delta^2} \log \left( \frac{C_2}{K \Delta^2} \right) + \frac{N C_1}{\Delta^2} \log T \\
&\ = \ 
N \frac{C_1}{\Delta^2} \Bigl[ \log \left( \frac{C_2}{K \Delta^2} \right) + \log T \Bigr] \\
&\ = \ 
N \frac{C_1}{\Delta^2} \log \left( \frac{C_2 T}{K \Delta^2} \right).
\end{align*}
Accounting for the constant $1 - NK$ on both sides yields,

\begin{align}
    \reG \leq N \frac{C_1}{\Delta^2} \log \left( \frac{C_2 T}{K \Delta^2} \right) + 1 - NK \in \bigO \left( \frac{N}{\Delta^2} \log \left( \frac{T}{K \Delta^2} \right)  \right)
\end{align}

\end{proof}

\clearpage

\subsection{$\mathsf{PrefLID}$ Algorithm} \label{app:preflid}

\begin{algorithm}[H]
\caption{Preference Resolution via Lattice Island Detection ($\mathsf{PrefLID}$)}
\label{alg:preflid}
\begin{algorithmic}[1]
\Require Agent sets $\seT{A}, \seT{B}$; computational budget $\rrBudget$;
         confidence level $\eta \in (0,1)$.
\Ensure Set of centroidal lattice matchings $\{\matcH^*_m\}$.
\State Initialize entrant set
       $\mathcal{U} \gets \{a_1, b_1, b_2\}$ with $a_1 \in \seT{A}$ and
       $b_1, b_2 \in \seT{B}$ arbitrary.
\State Initialize partial order $\parOrder$ as the completely uninformed
       ordering for every agent (one block per agent).
\State Initialize sampling counts to zero.
\While{$|\setlaT| > \rrBudget$ \textbf{or} $\setlaT = \emptyset$}
    \State Draw $c \sim \operatorname{Unif}((\seT{A} \cup \seT{B}) \setminus \mathcal{U})$.
    \State $\parOrder \gets \textsc{RRT}(c, \parOrder)$
           \Comment{$c$ performs a full round-robin against the opposite side.}
    \State Refresh Bradley--Terry estimates $\widehat{\theta}$ and
           recompute confidence intervals \eqref{eq:bt_ci}.
    \State Recompute partitions from confidence intervals and update
           $\parOrder$.
    \State $\setlaT \gets \emptyset$.
    \For{$\succ_{ab} \in \parOrder$}
        \State $\mathcal{L}_c \gets \textsf{LAT}(\succ_{ab})$.
        \State $\setlaT \gets \setlaT \cup \{\mathcal{L}_c\}$.
    \EndFor
    \If{$|\setlaT| < \rrBudget / |\seT{A} \cup \seT{B}|$}
        \State $\mathcal{U} \gets \mathcal{U} \cup \{c\}$.
    \EndIf
\EndWhile
\State $\isL \gets \mathsf{ConstructIslands}(\setlaT)$.
\If{$|\isL| = 1$}
    \State \Return $\{\matcH^*_1\}$.
\Else
    \State $\isL^* \gets \arg\max_{\isL^{(i)} \in \isL} |\isL^{(i)}|$
           \Comment{Select the largest island.}
    \State $\mathcal{Q} \gets \textsc{Enqueue}(\isL \setminus \isL^*)$
           \Comment{Queue remaining islands.}
    \While{$|\isL| > 1$}
        \State $\isL^{(i)} \gets \textsc{Dequeue}(\mathcal{Q})$.
        \State Select the unresolved pair $(a, b)$ with the widest
               confidence interval that lies within an unresolved
               partition of some agent in $\isL^{(i)}$.
        \State Sample $(a, b)$ until $\succ_{ab}$ is resolved with high
               probability.
        \If{$\succ_{ab}$ is disproved}
            \State Remove the corresponding lattice from $\setlaT$ and
                   refresh $\isL \gets \mathsf{ConstructIslands}(\setlaT)$.
        \EndIf
    \EndWhile
\EndIf
\State \Return $\{\matcH^*_m\}$.
\end{algorithmic}
\end{algorithm}

We now describe the \emph{Preference Resolution via Lattice Island
Detection} algorithm ($\mathsf{PrefLID}$). PrefLID is a two-sided,
explore-then-commit learner that certifies \emph{global} stability of a
committed matching by verifying that all configurations compatible with
the current uncertainty lead to a single lattice island. It improves on
the sample complexity of $\mathsf{P2ETG}$ (\S \ref{sec:p2etg}) by
certifying a matching at the level of refinements rather than at the
level of individual pairs.

\paragraph{Targeted Round-Robin (RRT) Sampling:}
Rather than sampling every pair $(b_i, b_j)$ for every agent $a$
uniformly, PrefLID centers each round on a single agent $c$ and has $c$
perform a full round-robin against the opposite side. Formally, if
$c \in \seT{A}$, the round consists of sampling the triplet
$(c; b_i, b_j)$ for every unordered pair $\{b_i, b_j\} \subseteq \seT{B}$,
yielding $K(K-1)/2$ comparisons in a single round; symmetrically for
$c \in \seT{B}$. Each round refines the partition structure of $c$ only,
since $c$'s confidence intervals are what govern $c$'s ordering; the
opposing agents' orderings are refined in their own rounds.

\paragraph{Stopping Condition:} We therefore introduce a stopping condition, where if $| \blockFunc(\setlaT)| = 1$, that is there exists only one island in the partitions of $ \blockFunc(\setlaT)$, then it must imply that the candidate matchings $\{ \matcH^{*}_m, \matcH^{*}_n, ... \matcH^{*}_l \}$ corresponding to the singleton island in $\blockFunc(\setlaT)$ exist as the true stable matchings, with probability $1 - \eta$. In other words, all rotations starting from various possible starting preference configurations (estimated under uncertainty) all share a common destination matching.

\clearpage

\section{Statistical Theory}

\subsection{Concentration Bound} \label{sec:deriv_delta_K}

Let there be \(K+1\) objects indexed by \(i\in\{0,\ldots,K\}\). For each unordered pair \(\{i,j\}\), suppose that \(T\) independent Bradley--Terry comparisons are observed, with,
\begin{align}
\mathbb{P}(i \succ j)
=
\frac{\theta_i}{\theta_i+\theta_j}.
\end{align}
Let
\begin{align}
a_i
:=
\sum_{j\neq i} a_{ij}
\end{align}
denote the total number of victories of object \(i\). Since object \(i\) is compared \(n\) times against each of the other \(K\) objects, \(a_i\) is a sum of \(NK\) independent Bernoulli random variables. Define the relative estimation error
\begin{align}
\Delta \theta_i
\deF
\frac{\hat{\theta}_i-\theta_i}{\theta_i},
\end{align}
and define the maximum merit ratio
\begin{align}
M_K
:=
\max_{0\leq i,j\leq K}
\frac{\theta_i}{\theta_j}.
\end{align}
% As in the standard Bradley--Terry normalization, set \(\mu_0=1\), so that \(\Delta \mu_0=0\).

\sectiondivider

For any fixed \(i\), and sample size (number of replications) $T$. Hoeffding's inequality gives,

\begin{align}
\mathbb{P}\left(
\left|a_i-\mathbb{E}[a_i]\right|>x
\right)
\leq
2\exp\left(
-\frac{2x^2}{TK}
\right).
\end{align}
Taking a union bound over all \(K+1\) objects yields
\begin{align}
\mathbb{P}\left(
\max_{0\leq i\leq K}
\left|a_i-\mathbb{E}[a_i]\right|>x
\right)
&\leq
\sum_{i=0}^{K}
\mathbb{P}\left(
\left|a_i-\mathbb{E}[a_i]\right|>x
\right) \\
&\leq
2(K+1)
\exp\left(
-\frac{2x^2}{TK}
\right).
\end{align}
To obtain a finite-sample concentration bound with prescribed failure probability \(\eta\in(0,1)\), choose \(x\) such that
\begin{align}
2(K+1)
\exp\left(
-\frac{2x^2}{TK}
\right)
=
\eta.
\end{align}
Solving this equation gives
\begin{align}
\exp\left(
-\frac{2x^2}{TK}
\right)
&=
\frac{\eta}{2(K+1)}, \\
-\frac{2x^2}{TK}
&=
\log\left(
\frac{\eta}{2(K+1)}
\right),
\end{align}
and hence
\begin{align}
x^2
=
\frac{TK}{2}
\log\left(
\frac{2(K+1)}{\eta}
\right).
\end{align}
Therefore,
\begin{align}
x
=
\sqrt{
\frac{TK}{2}
\log\left(
\frac{2(K+1)}{\eta}
\right)
}.
\end{align}
Thus,
\begin{align}
\mathbb{P}\left(
\max_{0\leq i\leq K}
\left|a_i-\mathbb{E}[a_i]\right|
\leq
\sqrt{
\frac{TK}{2}
\log\left(
\frac{2(K+1)}{\eta}
\right)
}
\right)
\geq
1-\eta.
\end{align}

It remains to translate this concentration inequality for the sufficient statistics \(a_i\) into a concentration inequality for the relative maximum likelihood errors \(\Delta \mu_i\). The deterministic comparison argument in \citep{Yao:1999_AsymptoticsWT} Lemma 3 implies that, whenever,
\begin{align}
\max_{0\leq i\leq K}
\left|a_i-\mathbb{E}[a_i]\right|
\leq r_K,
\end{align}
one obtains a bound of the form,
\begin{align}
\max_{0\leq i,j\leq K}
\left|
\Delta \mu_i-\Delta \mu_j
\right|
\leq
\frac{\delta_K}{1-\delta_K},
\end{align}
where
\begin{align}
\delta_K
=
8M_K\frac{r_K}{TK}.
\end{align}

\sectiondivider

% \lar{Requiring simplification/correction here.}

In the asymptotic statement of \citep{Yao:1999_AsymptoticsWT}, the radius is chosen as % \lar{check here.}
\begin{align}
r_K
=
\sqrt{TK\log(K+1)},
\end{align}
which gives
\begin{align}
\delta_K
=
8M_K
\frac{
\sqrt{TK\log(K+1)}
}{TK}
=
8M_K
\sqrt{
\frac{\log(K+1)}{TK}
}.
\end{align}

For a prescribed confidence level \(1-\eta\), we instead substitute the tunable Hoeffding radius % \lar{check here.}
\begin{align}
r_{K,\eta}
=
\sqrt{
\frac{TK}{2}
\log\left(
\frac{2(K+1)}{\eta}
\right)
}.
\end{align}
Therefore,
\begin{align}
\delta_{K,\eta}
&\deF
8M_K
\frac{r_{K,\eta}}{TK} \\
&=
8M_K
\frac{
\sqrt{
\frac{TK}{2}
\log\left(
\frac{2(K+1)}{\eta}
\right)
}
}{TK} \\
&=
8M_K
\sqrt{
\frac{1}{2TK}
\log\left(
\frac{2(K+1)}{\eta}
\right)
}.
\end{align}
Hence,
\begin{align}
\boxed{
\delta_{K,\eta}
=
8M_K
\sqrt{
\frac{1}{2TK}
\log\left(
\frac{2(K+1)}{\eta}
\right)
}
}.
\end{align}

\clearpage

\subsection{Estimation of Bradley-Terry Preference Parameters} \label{sec:BT_estimation}

\paragraph{Estimation:}
For a given agent $a \in \seT{A}$, let
$n^+_{a,(b_1,b_2)}$ and $n^-_{a,(b_1,b_2)}$ denote the number of observed
comparisons in which $a$ preferred $b_1$ over $b_2$ and $b_2$ over $b_1$
respectively. The Bradley--Terry log-likelihood for agent $a$ is
\begin{equation} \label{eq:bt_loglik}
    \mathcal{L}_a(\theta_a)
    = \sum_{b_1 < b_2}
      \Bigl[
        n^+_{a,(b_1,b_2)} \log
        \frac{\theta_{a,b_1}}{\theta_{a,b_1} + \theta_{a,b_2}}
        +
        n^-_{a,(b_1,b_2)} \log
        \frac{\theta_{a,b_2}}{\theta_{a,b_1} + \theta_{a,b_2}}
      \Bigr],
\end{equation}
and the maximum-likelihood estimate $\hat{\theta}_a$ is obtained by maximising
\eqref{eq:bt_loglik} subject to the normalisation in
Assumption~\ref{ass:reg_BT_params}. Symmetric expressions hold for each
$b \in \seT{B}$. Confidence intervals are constructed via a Hoeffding-type
bound: for each estimated coordinate $\hat{\theta}_{a,b}$ at time $t$, the
half-width takes the form
\begin{equation} \label{eq:bt_ci}
    w_{a,b}(t) = \sqrt{\frac{6 \log t_{a,b}}{T_{a,b}}},
\end{equation}
where $T_{a,b}$ is the number of comparisons contributing to
$\hat{\theta}_{a,b}$ and $t_{a,b}$ is the global time step at which the interval
is evaluated. % \lar{Confirm exact definition of $T_{a,b}$ and $t_{a,b}$.}

\sectiondivider

We employ the Bradley-Terry model, in which the probability that player $i$ beats player $j$ is
\[
P(i \succ j) = \frac{\theta_i}{\theta_i + \theta_j}, \qquad \theta_i > 0.
\]
Let $a_i = \sum_{j \neq i} w_{ij}$ denote the total number of wins of player $i$, and let $n_{ij} = w_{ij} + w_{ji}$ be the total number of observed comparisons between players $i$ and $j$.

The maximum likelihood estimators $\hat{\theta}_1,\dots,\hat{\theta}_n$ satisfy the system
\[
a_i = \sum_{j \neq i} \frac{n_{ij} \theta_i}{\theta_i + \theta_j}, \qquad i=1,\dots,n.
\]
These equations are solved using Zermelo's iterative method \citep{Zermelo:1929}.

\begin{algorithm}[H]
\caption{Iterative Estimation of Bradley--Terry Parameters (Zermelo's Iterative Method)}
\label{alg:bt_iterative_estimation}
\begin{algorithmic}[1]
\Require Pairwise comparison counts $\{n_{ij}\}_{i,j=1}^n$, win counts $\{a_i\}_{i=1}^n$, tolerance $\varepsilon > 0$
\Ensure Estimated parameters $\{\widehat{\theta}_i\}_{i=1}^n$

\State Initialize $\theta_i^{(0)} > 0$ for all $i \in \{1,\dots,n\}$, e.g.,
\[
\theta_i^{(0)} = 1
\quad \text{or} \quad
\theta_i^{(0)} = a_i + 1.
\]

\State Set $k \gets 0$

\Repeat
    \For{$i = 1,\dots,n$}
        \State Update
        \[
        \theta_i^{(k+1)}
        =
        \frac{a_i}{
        \sum_{j \neq i}
        \frac{n_{ij}}{\theta_i^{(k)} + \theta_j^{(k)}}
        }.
        \]
    \EndFor

    \State Compute the relative change
    \[
    \delta^{(k)}
    =
    \max_{i}
    \left|
    \frac{\theta_i^{(k+1)} - \theta_i^{(k)}}{\theta_i^{(k)}}
    \right|.
    \]

    \State Set $k \gets k+1$

\Until{$\delta^{(k-1)} < \varepsilon$}

\State \Return $\widehat{\theta}_i = \theta_i^{(k)}$ for all $i \in \{1,\dots,n\}$

\end{algorithmic}
\end{algorithm}

After convergence, the estimates may be normalized (e.g., by setting $\hat{\theta}_1 = 1$ or $\sum_i \hat{\theta}_i = 1$) for identifiability and interpretability. The procedure converges reliably provided the comparison graph is connected. This method yields the maximum likelihood estimates of the preference parameters $\theta_i$.

\clearpage

\subsection{A Sample Complexity Result for P2ETC-GS}

We ask the question of how many samples does it to require, such that with high probability $1- \eta$, we can obtain an estimate of parameter gap $\hat{\theta}_i -\hat{\theta}_j$ that corresponds with the true parameter gap $\theta_i -\theta_j$. This correspondence allows to build a ranking over preferences $\succ_a$ and $\succ_b$ respectively. From \citep{Yao:1999_AsymptoticsWT} we have the result,

% , i.e. $\hat{\theta}_i -\hat{\theta}_j > 0 \implies \theta_i -\theta_j$.

% Provide a sample complexity result for ETC to produce w.h.p. Gale Shapley result, we look at a concentration bound for pair-wise comparisons,

\begin{align}
    \max_{i,j \in \setA} \left| \frac{\hat{\theta}_i - \theta_i}{\theta_i} - \frac{\hat{\theta}_j - \theta_j}{\theta_j} \right| 
    \leq \frac{\delta_{K}}{1 - \delta_{K}}   
    %= \frac{\delta_{K, \eta}}{\sqrt{T_{ij}} - \delta_{K, \eta}} 
    \in \bigO\left( \frac{1}{\sqrt{T_{ij}}} \right).
\end{align}

Where from \citep{Yao:1999_AsymptoticsWT} $\delta_{K}$ serves as a scaling function,

% \begin{align}
%     \delta_{K, \eta} = 8 M_K \sqrt{\frac{\log(K+1)}{K}}, \qquad M_K = \max_{i,j \in \setA} \frac{\theta_i}{\theta_j}.
% \end{align}

% \begin{align}
%     \delta_{K, \eta}
%         \deF
%         8M_K
%         \sqrt{
%         \frac{1}{2TK}
%         \log\left(
%         \frac{2(K+1)}{\eta}
%         \right)}, \qquad M_K = \max_{i,j \in \setA} \frac{\theta_i}{\theta_j}.
% \end{align}

\begin{align}
    \delta_{K} \deF 8M_K \sqrt{\frac{\log(K+1)}{TK}}, 
        \qquad M_K = \max_{i,j \in \setA} \frac{\theta_i}{\theta_j}.
\end{align}

Where,

\begin{align}
M_K = o\!\left(\sqrt{\frac{K}{\log K}}\right) \quad \text{as } t \to \infty,
\qquad \text{i.e.,} \quad 
\lim_{K \to \infty} \frac{M_K}{\sqrt{K / \log K}} = 0.
\end{align}

The key idea is that we must draw a sufficient amount of samples for between any proposer $i \in [N]$ and $K$ receivers, each pair ($i$, $j$) must be sampled $T$ times, such that we obtain high confidence estimate of the preference relation $\tT{Ord}(\theta_i)$. (Please see \S \ref{sec:deriv_delta_K} for derivation of $\delta_{K, \eta}$.)

\begin{theorem} \label{thm:gs_sample_complexity}
    \textbf{Sample Complexity for GS:} For $i \in [N]$  proposers, and $j \in [K]$ receivers, each pair ($i$, $j$) the sample complexity such that number of pairwise comparison between $K$ receivers, produces with probability, $1-\eta$ a correct ordering when $T$ samples are obtained,
     
    \[
    T \in \bigO \left(  \frac{M_K^2(1+\Delta)^2}
        {\Delta^2K}
        \log\!\left(
        \frac{2(K+1)}{\eta}
        \right) \right). 
            \]
    where, $\Delta \deF \min_{i,j \in \setA} \left| 1/\theta_i - 1/\theta_j  \right|$.
\end{theorem}

\begin{proof}

    For any agent $i$ pairwise against all $K$ opponents,

    \[
    \Pr\!\left(
    \max_{0\le i\le K}
    |a_i-\mathbb E[a_i]|
    \ge x
    \right)
    \le
    \exp\!\left(
    -\frac{2x^2}{TK}
    \right).
    \]

The result is obtained from Hoeffding's inequality together with a union bound. Thus for $TK$ independent Bernoulli trials, for every $x>0$,
\[
\Pr\!\left(
\max_{0\le i\le K}
|a_i-\mathbb E[a_i]|
\ge x
\right)
\le
2(K+1)\exp\!\left(
-\frac{2x^2}{TK}
\right).
\]
Setting the right-hand side equal to $\eta$ yields
\[
x
=
\sqrt{
\frac{TK}{2}
\log\!\left(
\frac{2(K+1)}{\eta}
\right)
}.
\]
Hence,
\[
\Pr\!\left(
\max_i
|a_i-\mathbb E[a_i]|
<
\sqrt{
\frac{TK}{2}
\log\!\left(
\frac{2(K+1)}{\eta}
\right)
}
\right)
\ge
1-\eta.
\]

The proof of Lemma 3 from \citep{Yao:1999_AsymptoticsWT} is purely deterministic once the above event holds.
Replacing the threshold
\[
\sqrt{TK\log(K+1)}
\]
by
\[
\sqrt{
\frac{TK}{2}
\log\!\left(
\frac{2(K+1)}{\eta}
\right)
}
\]
throughout the proof gives,
\[
1-\frac{\beta}{\alpha}
<
8M_K
\sqrt{
\frac{\log\!\left(\frac{2(K+1)}{\eta}\right)}
{2TK}
}
\deF \delta_\eta,
\]
where
\[
\alpha=\max_i\frac{\hat \theta_i}{\theta_i},
\qquad
\beta=\min_i\frac{\hat \theta_i}{\theta_i}.
\]
Exactly as in Lemma 3 \citep{Yao:1999_AsymptoticsWT},
\[
\max_{i,j}
|\Delta_i-\Delta_j|
=
\max_{i,j}
\left|
\frac{\hat \theta_i}{\theta_i}
-
\frac{\hat \theta_j}{\theta_j}
\right|
\le
\frac{\alpha-\beta}{\beta}
<
\frac{\delta_\eta}{1-\delta_\eta}.
\]
Therefore,
\[
\Pr\!\left(
\max_{i,j}
|\Delta_i-\Delta_j|
<
\frac{\delta_\eta}{1-\delta_\eta}
\right)
\ge
1-\eta.
\]

Finally, to guarantee an estimation error at most $\Delta$, it is sufficient
that
\[
\frac{\delta_\eta}{1-\delta_\eta}
\le
\Delta,
\]
which is equivalent to
\[
\delta_\eta
\le
\frac{\Delta}{1+\Delta}.
\]

Specifically, for the gap $\Delta$, we note that without loss of generality, for any $i, j$,

\[
|\Delta_i-\Delta_j| \leq  \Delta \leq \left| \frac{1}{\theta_i} - \frac{1}{\theta_j} \right|,
\]

since via Ass. \ref{ass:reg_BT_params} $\sum_{i=1}^N \theta_i = 1$. Substituting the expression for $\delta_\eta$ gives,
\[
8M_K
\sqrt{
\frac{\log\!\left(\frac{2(K+1)}{\eta}\right)}
{2TK}
}
\le
\frac{\Delta}{1+\Delta}.
\]
Squaring both sides and rearranging yields
\[
T
\ge
\frac{32M_K^2(1+\Delta)^2}
{\Delta^2K}
\log\!\left(
\frac{2(K+1)}{\eta}
\right),
\]
which proves the claim, where $\theta \deF \min_{i,j} \left| \frac{1}{\theta_i} - \frac{1}{\theta_j} \right| $ is minimised such that $T$ is conservatively maximised.
    
\end{proof}

\section{Auxilliary Lemmas and Corollaries}

\begin{lemma}\label{lem:factorial_diff}
For any integers \( a, b \in \mathbb{N} \) with \( a \geq b+2 \), we have
\[
(a-1)! + (b+1)! < a! + b!.
\]
\end{lemma}

\begin{proof}
Compute:
\[
(a! + b!) - ((a-1)! + (b+1)!) = (a-1)(a-1)! - b \cdot b!.
\]
Since \( a \geq b+2 \), we have \( a-1 \geq b+1 \), so
\[
(a-1)(a-1)! \geq (b+1)(b+1)! > b \cdot b!.
\]
Thus the difference is positive, proving the lemma.
\end{proof}

% \subsection{Proof of Lemma \ref{lem:balanced_partition_mini}} \label{prf:balanced_partition_mini}

\begin{lemma}\label{lem:balanced_partition_mini}
    \TXTBalancedPartitionMini
\end{lemma}

% (Proof provided in Appendix \ref{prf:balanced_partition_mini}.)

% \noindent \textit{Sketch of Proof:} The lemma proves that for \(a \geq b+2\), replacing \(a!+b!\) with \((a-1)!+(b+1)!\) reduces the sum, as the difference simplifies to \((a-1)(a-1)! - b\cdot b! > 0\).  In the main proof, this inequality is used to show that any minimizing partition cannot have two subset sizes differing by more than 1, since moving an element from a larger to a smaller subset would lower the cost.  

\begin{proof}
Suppose a minimizer has two subsets of sizes \( a \) and \( b \) with \( a \geq b+2 \). 
By Lemma \ref{lem:factorial_diff}, moving one element from the larger to the smaller subset strictly decreases 
the total cost, contradicting minimality. Hence all sizes differ by at most 1. 
The unique integer solution to \( \sum n_i = N \) with \( \max n_i - \min n_i \leq 1 \) 
is exactly the one given, yielding the stated minimum cost. Therefore, the optimal partition must have all subset sizes as equal as possible, and the unique balanced distribution gives the minimum total cost.

\end{proof}

\begin{lemma} \label{label:poly_Nd_scaling}
    \textbf{Fixed Parameter Polynomial Growth of $|\widehat{\succ}|$ via Minimising Cuts:} When cuts $C_\Psi$ are placed optimally, minimising the size of $|\widehat{\succ}|$, it is possible to construct a algorithm that scales polynomially with respect to $\bar{N}$, for the enumeration over all refinements of $\widehat{\succ}$.
\end{lemma}

% find a fixed parameter tractable algorithm (FPT) (slicewise polynomial)

\begin{proof}
We choose \(C_\Psi\) to grow polynomially with \(N\), namely $C_\Psi = N^d$, then,
\[
\frac{N}{C_\Psi+1}+1 = \frac{N}{N^d+1}+1.
\]
For $d \ge 1$ and $N$ sufficiently large (indeed for all $N \ge 1$ when $d=1$),
\[
\frac{N}{N^d+1} < 1,
\]
hence,
\[
\frac{N}{N^d+1}+1 < 2.
\]
Thus, the factorial argument is strictly less than $2$, so its factorial is bounded by a constant (at most $2$). Therefore,
\[
C_\Psi\left(\frac{N}{C_\Psi+1}+1\right)!
\le N^d \cdot 2 = O(N^d).
\]
For $d=1$, this gives the sharper bound $2N$. % \lar{FPT Argument.}

\end{proof}

\clearpage

\section{Supplemental Algorithms} \label{app:supp_alg}

\begin{algorithm} 
\caption{Gale-Shapley Deferred Acceptance Algorithm \citep{Gale:1962}} \label{alg:gale_shapley}
\begin{algorithmic}[1] 
\Require Sets $A$ and $B$ with strict preference lists $\succ_a$ for each $a \in A$ and $\succ_b$ for each $b \in B$
\Ensure A stable matching $M \subseteq A \times B$
\State Initially, all agents are free (unmatched).
\While{some agent $a \in A$ remains free and has not proposed to every $b \in B$}
    \State Let $b$ be the highest-ranked agent in $B$ that $a$ has not yet proposed to.
    \State $a$ proposes to $b$.
    \If{$b$ is free}
        \State $a$ and $b$ become engaged.
    \Else
        \State $b$ compares $a$ and $a'$ (where $a'$ is $b$'s current partner) according to $\succ_b$.
        \If{$a \succ_b a'$}
            \State $b$ breaks the engagement with $a'$ and becomes engaged to $a$.
            \State $a'$ becomes free.
        \Else
            \State $a$ remains free.
        \EndIf
    \EndIf
\EndWhile
\State \Return the set of engaged pairs.
\end{algorithmic}
\end{algorithm}

\begin{algorithm}[h!]
\caption{Island Construction via Centroidal Lattice Matchings}
\label{alg:island_construction}
\begin{algorithmic}[1]
\Require Set of lattices $\setlaT$ from preference configurations $\parOrder$.
\Ensure Island decomposition $\isL = \blockFunc(\setlaT)$, one centroidal matching $\matcH^*_m$ per island.
\State Initialize $\isL \gets \emptyset$, $m \gets 0$.
\While{$\setlaT \neq \emptyset$}
    \State $m \gets m{+}1$; pick seed $\laT^{(i)} \in \setlaT$; set $\isL^{(m)} \gets \{\laT^{(i)}\}$; $\matcH^*_m \gets \text{any } \matcH \in V(\laT^{(i)})$.
    \State Add every $\laT^{(k)} \in \setlaT$ with $\matcH^*_m \in V(\laT^{(k)})$ to $\isL^{(m)}$.
    \State Update $\matcH^*_m$ to the matching shared by the largest number of lattices in $\isL^{(m)}$.
    \State Add $\isL^{(m)}$ to $\isL$; remove its lattices from $\setlaT$.
\EndWhile
\State \Return $\isL = \blockFunc(\setlaT)$.
\end{algorithmic}
\end{algorithm}

\begin{algorithm}[H]
\caption{Targeted Round-Robin (RRT) for a Single Center Agent}
\label{alg:rrt_round}
\begin{algorithmic}[1]
\Require Center agent $c$; agent states $\{\widehat{\theta}_a, n_a\}$.
\Ensure Updated counts $n_c$; global time advanced by $C(|\mathcal{O}|, 2)$.

\State $\mathcal{O} \gets \text{opposite side of } c$
\ForAll{unordered pairs $\{b_i, b_j\} \subseteq \mathcal{O}$,
        in ascending order of $n_c[\operatorname{can}(b_i, b_j)]$}
    \State $X \sim \operatorname{Bernoulli}\!\left(
              \dfrac{\widehat{\theta}_{c,b_i}}
                    {\widehat{\theta}_{c,b_i} + \widehat{\theta}_{c,b_j}}
           \right)$
    \State $n_c.\mathrm{record}(b_i, b_j, X)$; \quad $t \gets t + 1$
\EndFor
\end{algorithmic}
\end{algorithm}

\clearpage

\section{Empirical Study} \label{app:empirical}

\begin{figure*}[!h]
\centering
\subfloat[Scenario: $N=K$.]{%
  \includegraphics[width=49mm]{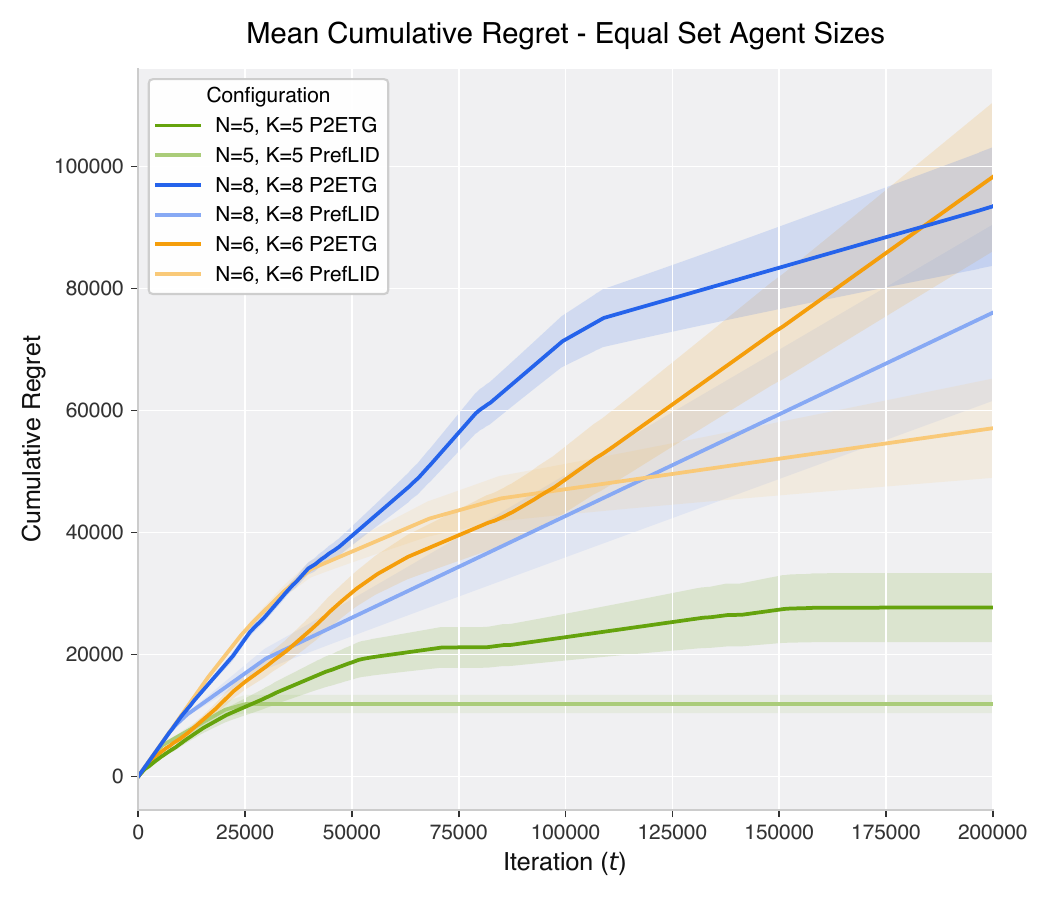}%
  \label{fig:reg_equal_nk}%
}\hspace{10pt}% Add gap here
\subfloat[Scenario: $N\neq K$.]{%
  \includegraphics[width=49mm]{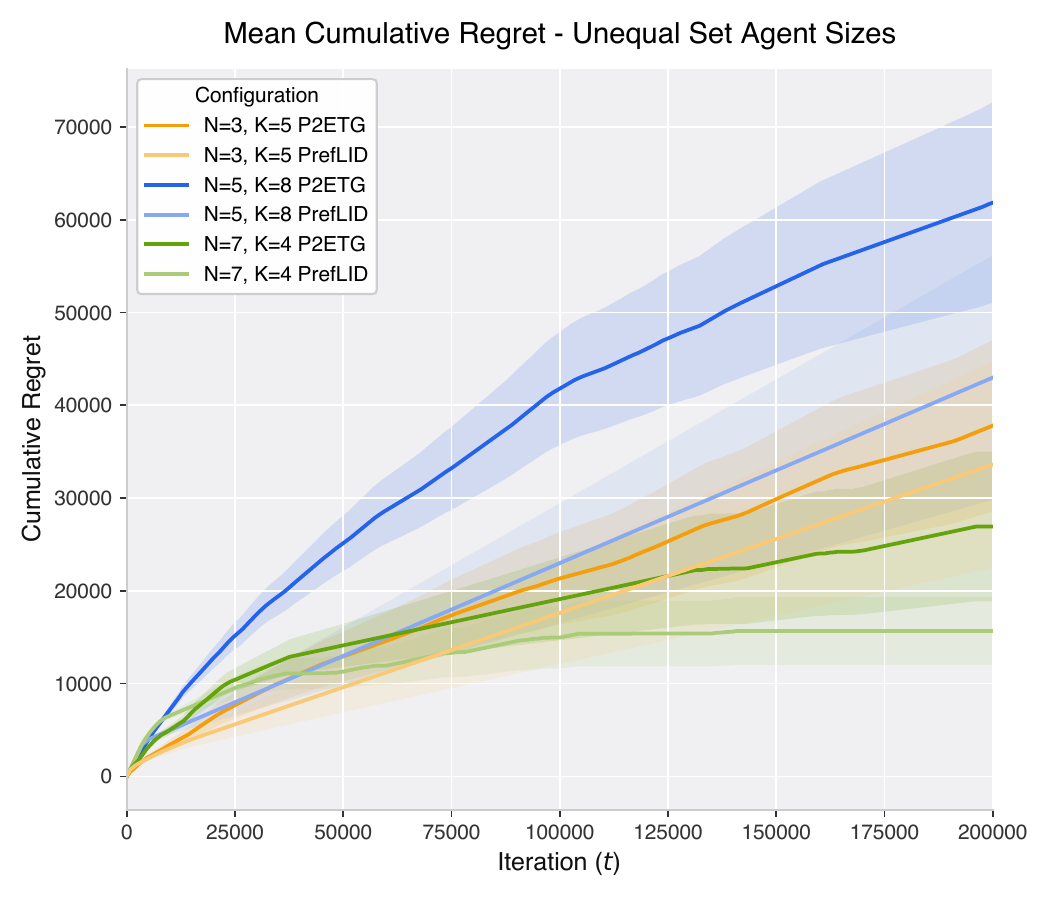}%
  \label{fig:reg_unequal_nk}%
}\hspace{10pt}% Add gap here
\subfloat[Scenario: LLM Routing.]{%
  \includegraphics[width=49mm]{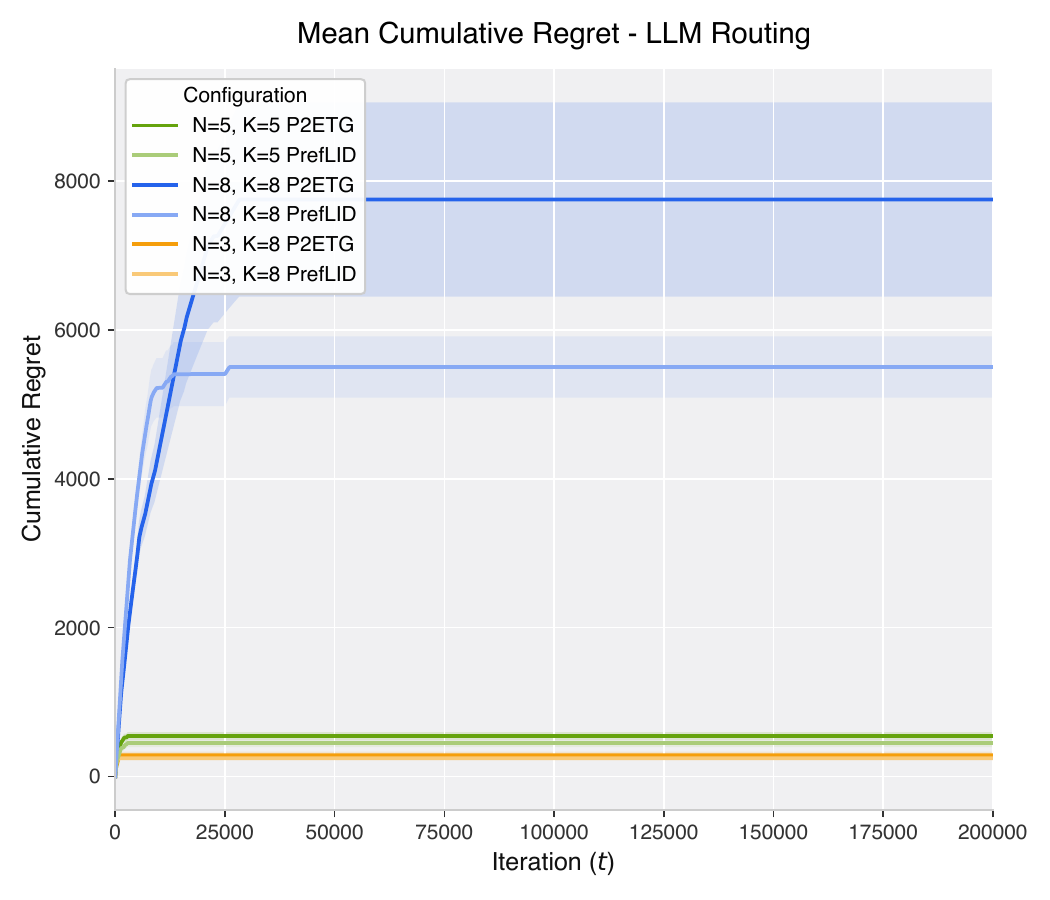}
  \label{fig:reg_llm}
}
\caption{
Ablation Study of Cumulative Regret for Various Stable Matching Environments. 
} \label{fig:all_regret_plots_main}
% \vspace{-15pt}
\end{figure*}

\begin{table}[htbp]
\centering
\small
\begin{tabular}{c l l c c r S[table-format=1.4e1] S[table-format=1.4] S[table-format=1.4] S[table-format=1.3]}
\toprule
\textbf{Pair} & \textbf{Data} & \textbf{ALG} & \textbf{N} & \textbf{K} & \textbf{$T_{\text{stop}}$} & \textbf{Final Regret} & \textbf{$|\widetilde{\seT{M}}|$} & $\Delta$ & \textbf{$|\isL|$} \\
\midrule
\multicolumn{10}{l}{\textit{Group 1: $N = K$ (Synthetic Data)}} \\
\midrule
% ---- (N=5, K=5) ----
1 & Synthetic Data & P2ETG   & 5  & 5  & \num{1.562e5} & 2.769e4 & 1.400 & \num{2.137e-3} & {-} \\
1 & Synthetic Data & PrefLID & 5  & 5  & \num{1.094e5} & 1.188e4 & 1.800 & \num{3.482e-3} & 1.284 \\
% ---- (N=6, K=6) ----
2 & Synthetic Data & P2ETG   & 6  & 6  & \num{1.730e5} & 9.830e4 & 2.300 & \num{1.264e-3} & {-} \\
2 & Synthetic Data & PrefLID & 6  & 6  & \num{1.211e5} & 5.709e4 & 2.700 & \num{3.915e-3} & 2.067 \\
% ---- (N=8, K=8) ----
3 & Synthetic Data & P2ETG   & 8  & 8  & \num{1.850e5} & 9.347e4 & 2.054 & \num{2.806e-3} & {-} \\
3 & Synthetic Data & PrefLID & 8  & 8  & \num{1.295e5} & 7.603e4 & 2.833 & \num{1.549e-3} & 3.142 \\
\midrule
\multicolumn{10}{l}{\textit{Group 2: $N \neq K$ (Synthetic Data)}} \\
\midrule
% ---- (N=3, K=5) ----
1 & Synthetic Data & P2ETG   & 3  & 5  & \num{1.680e5} & 3.782e4 & 1.083 & \num{3.673e-3} & {-} \\
1 & Synthetic Data & PrefLID & 3  & 5  & \num{1.176e5} & 3.362e4 & 1.000 & \num{2.294e-3} & 1.073 \\
% ---- (N=5, K=8) ----
2 & Synthetic Data & P2ETG   & 5  & 8  & \num{1.750e5} & 6.184e4 & 1.056 & \num{1.018e-3} & {-} \\
2 & Synthetic Data & PrefLID & 5  & 8  & \num{1.225e5} & 4.299e4 & 1.288 & \num{3.841e-3} & 2.315 \\
% ---- (N=7, K=4) ----
3 & Synthetic Data & P2ETG   & 7  & 4  & \num{1.900e5} & 2.693e4 & 1.311 & \num{2.562e-3} & {-} \\
3 & Synthetic Data & PrefLID & 7  & 4  & \num{1.330e5} & 1.567e4 & 1.100 & \num{1.437e-3} & 2.748 \\
\midrule
\multicolumn{10}{l}{\textit{Group 3: LLM Routing}} \\
\midrule
% ---- (N=3, K=8) ----
1 & LLM Routing & P2ETG   & 3  & 8  & \num{8.940e2} & 2.852e2 & 1.000 & \num{3.205e-3} & {-} \\
1 & LLM Routing & PrefLID & 3  & 8  & \num{1.910e2} & 2.449e2 & 1.000 & \num{1.872e-3} & 1.512 \\
% ---- (N=5, K=5) ----
2 & LLM Routing & P2ETG   & 5  & 5  & \num{1.320e4} & 5.444e2 & 1.000 & \num{2.941e-3} & {-} \\
2 & LLM Routing & PrefLID & 5  & 5  & \num{1.119e4} & 4.480e2 & 1.000 & \num{1.336e-3} & 1.948 \\
% ---- (N=8, K=8) ----
3 & LLM Routing & P2ETG   & 8  & 8  & \num{1.017e5} & 7.750e3 & 1.000 & \num{3.774e-3} & {-} \\
3 & LLM Routing & PrefLID & 8  & 8  & \num{1.124e5} & 5.499e3 & 1.000 & \num{2.018e-3} & 3.276 \\
\bottomrule
\end{tabular}
\caption{Performance summary across configurations, grouped into $N=K$ (synthetic), $N\neq K$ (synthetic), and LLM routing. Each pair compares P2ETG vs.\ PrefLID at various configurations for $N$ and $K$. All simulations were run to $200,000$ iterations, at 10 replications each. In addition the average number of stable matchings $\widetilde{\seT{M}}$, average minimum gap of Bradley Terry parameters $\Delta$, and average number of islands in the lattice island set $|\isL|$ is recorded. The budget parameter $\rrBudget = 2 \times 10^4$.}
\label{tab:performance-summary}
\end{table}

\subsection{Experiments on Real LLM Routing Markets} \label{sec:llm_experiments}

We instantiate the framework on real evaluation data and evaluate $\mathsf{P2ETG}$ and $\mathsf{PrefLID}$ end to end. No LLM inference is performed at any point: the environment is built once from stored benchmark scores, and every pairwise comparison the learners request is a Bernoulli draw against this fixed ground truth.

\paragraph{Environments:}
We use the LLMRouterBench evaluation suite~\cite{li2026llmrouterbench} (20 models $\times$ 15 task types, per-instance scores). From the full pool we greedily select markets of size $N \in \{3, 5, 8, 10\}$: tasks are added in decreasing order of their full-pool winning margin, together with their distinct top models (an $8 \times 8$ market requires two runner-up models to complete both sides). Task-side utilities are mean scores; model-side preferences are the transpose of the same matrix, giving exactly the reciprocal structure above: a unique stable matching $\SMSet$, verified at construction (men- and women-proposing Gale--Shapley coincide). Feedback follows the Bradley--Terry model of Section~\ref{sec:bradley_terry}. Exact utility ties --- which would produce coin-flip comparisons no algorithm can resolve --- are removed by a \emph{probability floor} of $0.15$: every win probability is clamped to $|p - \tfrac12| \ge 0.15$ in the direction of the true preference. This changes only the observation channel; preferences and $\SMSet$ are untouched. Regret is the identification regret of Eq.~\eqref{eq:reg_def}.

\paragraph{Configuration:}
Both learners use Hoeffding confidence intervals and certified stopping. $\mathsf{P2ETG}$ stops when all pairwise confidence intervals exclude $\tfrac12$; $\mathsf{PrefLID}$ stops on lattice-island certification with a round-robin centre schedule. The sound constant $c = 0.3$ is used for $\mathsf{PrefLID}$'s certificate at $5\times5$ and $8\times8$ (smaller constants false-certify at low sample counts); full hyperparameters appear in Appendix~\ref{app:empirical}. Each cell reports 10 seeds, horizon $2 \times 10^5$ comparisons, $\texttt{PYTHONHASHSEED}=0$.

\paragraph{Results:}
Table~\ref{tab:performance-summary} shows three findings. \emph{(i) Real markets are learnable to perfection.} Across all four sizes, both algorithms recover the unique stable matching in every run: all 40 committed and final matchings equal $\SMSet$, and every stop is correct. \emph{(ii) Certification scales better than full resolution.} At $3\times3$ the structural certificate fires $4.6\times$ earlier than full preference resolution (189 vs.\ 864 comparisons). The gap closes at intermediate sizes, where the sound constant prices the certificate at a constant factor. At $10\times10$ the picture inverts decisively: $\mathsf{P2ETG}$'s stopping condition requires all $900$ pairwise comparisons resolved \emph{simultaneously}, and only 3 of 10 seeds manage within the horizon, whereas $\mathsf{PrefLID}$ certifies in 9 of 10 seeds at a median of $1.2\times10^5$ comparisons --- identifying the assignment does not require fully resolving every preference. \emph{(iii) Correctness is a hyperparameter, not luck.} With the upstream constant $c=0.01$, certificates fire at ${\approx}25$ samples per arm and commit wrong matchings in $80\%$ of runs; with $c=0.1$, the $5\times5$ market still yields one false certificate out of ten. Only the sound constant $c=0.3$ certifies correctly in all 30 validation seeds. The no-tie reward floor is the complementary lever: without it, the $10\times10$ market contains near-ties whose comparisons are effectively coin flips, and no amount of sampling certifies the matching within the horizon.

\subsection{Data Source and Market Construction}

\paragraph{Source:}
The environments are built from the LLMRouterBench evaluation suite~\cite{li2026llmrouterbench}
(20 models $\times$ 15 task types; one stored per-instance score, cost,
and token count per (model, task, instance); no LLM inference is run).
Records are aligned per task by intersecting instance indices across
the selected models; any missing entry is a hard error, and a quality
gate requires at least 100 aligned records per task. Utilities are
computed on the \emph{full} aligned data (no train/test split --- the
online setting has a fixed ground truth).

\paragraph{Market selection:}
From the full $15 \times 20$ utility matrix, tasks are greedily added in
decreasing order of full-pool winning margin (winner minus runner-up),
together with their distinct top models. Sizes $3$, $5$, and $10$ are
prefixes of this greedy order; the $8\times8$ market is the diverse
flagship selection of the same pool. The $10\times10$ market completes
both sides with the two runner-up models of the remaining
highest-margin tasks.

\begin{table}[h!]
\centering
\caption{Composition of the four routing markets of Table~\ref{tab:performance-summary}:
models ($\seT{A}$-side) and tasks ($\seT{B}$-side). The $8\times8$ market is
the diverse flagship selection of the same pool.}
\label{tab:market_composition}
\small
\begin{tabular}{@{}p{0.11\linewidth} p{0.42\linewidth} p{0.39\linewidth}@{}}
\toprule
Market & Models ($\seT{A}$) & Tasks ($\seT{B}$) \\
\midrule
$3\times3$ &
\textit{Qwen2.5-Coder-7B-Instruct}, \textit{Qwen3-8B},
\textit{DeepSeek-R1-0528-Qwen3-8B} &
\textit{mbpp}, \textit{finqa}, \textit{gpqa} \\
\midrule
$5\times5$ &
\textit{Qwen2.5-Coder-7B-Instruct}, \textit{Qwen3-8B},
\textit{DeepSeek-R1-0528-Qwen3-8B}, \textit{Intern-S1-mini},
\textit{gemma-2-9b-it} &
\textit{mbpp}, \textit{finqa}, \textit{gpqa}, \textit{kandk},
\textit{emorynlp} \\
\midrule
$8\times8$ &
\textit{DeepSeek-R1-0528-Qwen3-8B}, \textit{Fin-R1},
\textit{GLM-Z1-9B-0414}, \textit{Intern-S1-mini},
\textit{Qwen2.5-Coder-7B-Instruct}, \textit{Qwen3-8B},
\textit{gemma-2-9b-it}, \textit{glm-4-9b-chat} &
\textit{emorynlp}, \textit{finqa}, \textit{humaneval}, \textit{korbench},
\textit{math500}, \textit{mbpp}, \textit{medqa}, \textit{meld} \\
\midrule
$10\times10$ &
\textit{Qwen2.5-Coder-7B-Instruct}, \textit{Qwen3-8B},
\textit{DeepSeek-R1-0528-Qwen3-8B}, \textit{Intern-S1-mini},
\textit{gemma-2-9b-it}, \textit{GLM-Z1-9B-0414}, \textit{glm-4-9b-chat},
\textit{MiniCPM4.1-8B}, \textit{NVIDIA-Nemotron-Nano-9B-v2}, \textit{Fin-R1} &
\textit{mbpp}, \textit{finqa}, \textit{gpqa}, \textit{kandk},
\textit{emorynlp}, \textit{math500}, \textit{meld}, \textit{mathbench},
\textit{livecodebench}, \textit{humaneval} \\
\bottomrule
\end{tabular}
\end{table}

\paragraph{Preferences and oracle:}
Task-side utility is the mean score; model-side preference is the
transpose of the same matrix (reciprocal structure). Exact ties within
$10^{-6}$ receive a deterministic SHA-256 jitter that preserves the
reciprocal structure. Under reciprocity the stable matching is unique;
construction verifies that men-proposing and women-proposing
Gale--Shapley coincide, and both equal the mutual-best cascade.

\subsection{Feedback Model}

\paragraph{Probability floor (no-tie guarantee):}
Every win probability is clamped to $|p - \tfrac12| \ge \varepsilon_{\mathrm{floor}}$ with $\varepsilon_{\mathrm{floor}} = 0.15$, in the direction of the (jittered) true preference: exact ties become $p = 0.5 \pm \varepsilon_{\mathrm{floor}}$; weak signals below the floor are boosted to the floor; stronger signals are untouched. The floor changes only the observation channel --- preferences, oracle, and welfare are unchanged --- and guarantees that every pairwise arm resolves within $n \approx c\ln(t)/\varepsilon_{\mathrm{floor}}^2$ samples.

\clearpage

\end{document}